\documentclass[
    twocolumn,
    byrevtex,
    superscriptaddress,
    floatfix,
    nofootinbib,
    longbibliography,
]{revtex4-1}
\setcitestyle{authoryear, round}
\usepackage{xr}
\usepackage[realmainfile]{currfile}
\usepackage{color}

\usepackage{graphicx,epsfig,verbatim,enumerate}
\usepackage{amssymb,amsmath}
\usepackage{ifthen}

\PassOptionsToPackage{hyphens}{url}\usepackage{hyperref}
\makeatletter
\g@addto@macro{\UrlBreaks}{\UrlOrds}
\makeatother

\usepackage{hyperref}
\hypersetup{
  colorlinks=true,
  allcolors=black,
  urlcolor=black,
  citecolor=RoyalBlue,
  pdfborder={0 0 0},
  breaklinks=true,
}

\usepackage{mathtools}

\usepackage{tabularx}
\usepackage{multirow, makecell}
\newcolumntype{C}{>{\centering\arraybackslash}X}
\newcolumntype{L}{>{\raggedright\arraybackslash}X}
\newcolumntype{R}{>{\raggedleft\arraybackslash}X}
\usepackage[english]{babel}
\usepackage{natbib}

\usepackage[latin1,utf8]{inputenc}
\usepackage[normalem]{ulem}

\usepackage{subfigure}
\usepackage{longtable}

\usepackage{placeins}
\usepackage{booktabs}

\newboolean{twocolswitch}

\newcommand{\sindex}[1]{}
\newcommand{\nindex}[1]{}

\newcommand{\etal}{\textit{et al.}}
\newcommand{\www}[1]{\url{#1}}

\usepackage{lettrine}

\usepackage[dvipsnames]{xcolor}

\usepackage{xifthen}
\newcommand{\dummycite}[2][]{\ifthenelse{\isempty{#2}}{\textcolor{NavyBlue}{[CITE]}}{\textcolor{NavyBlue}{[CITE\textemdash #2]}}}

\setboolean{twocolswitch}{true}

\begin{document}

\title{\protect
A decomposition of book structure through ousiometric fluctuations in cumulative word-time
}

\author{
\firstname{Mikaela Irene}
\surname{Fudolig}
}
\email{mikaela.fudolig@uvm.edu}

\affiliation{
  Computational Story Lab,
  Vermont Complex Systems Center,
  MassMutual Center of Excellence for Complex Systems and Data Science,
  Vermont Advanced Computing Core,
  University of Vermont,
  Burlington, VT, USA
  }

\author{
\firstname{Thayer}
\surname{Alshaabi}
}

 \affiliation{
  Advanced Bioimaging Center,
  UC Berkeley,
  Berkeley, CA, USA
  }
  
\affiliation{
  Computational Story Lab,
  Vermont Complex Systems Center,
  MassMutual Center of Excellence for Complex Systems and Data Science,
  Vermont Advanced Computing Core,
  University of Vermont,
  Burlington, VT, USA
  }

\author{
\firstname{Kathryn}
\surname{Cramer}
}

\affiliation{
  Computational Story Lab,
  Vermont Complex Systems Center,
  MassMutual Center of Excellence for Complex Systems and Data Science,
  Vermont Advanced Computing Core,
  University of Vermont,
  Burlington, VT, USA
  }

\author{
\firstname{Christopher M.}
\surname{Danforth}
}

\affiliation{
  Computational Story Lab,
  Vermont Complex Systems Center,
  MassMutual Center of Excellence for Complex Systems and Data Science,
  Vermont Advanced Computing Core,
  University of Vermont,
  Burlington, VT, USA
  }

\affiliation{
  Department of Mathematics \& Statistics,
  University of Vermont,
  Burlington, VT, USA
  }

\author{
\firstname{Peter Sheridan}
\surname{Dodds}
}

\affiliation{
  Computational Story Lab,
  Vermont Complex Systems Center,
  MassMutual Center of Excellence for Complex Systems and Data Science,
  Vermont Advanced Computing Core,
  University of Vermont,
  Burlington, VT, USA
  }

\affiliation{
  Department of Computer Science,
  University of Vermont,
  Burlington, VT, USA
}

\begin{abstract}
  \protect
While quantitative methods have been used to examine changes in word usage in books, studies have focused on overall trends, such as the shapes of narratives, which are independent of book length.
We instead look at how words change over the course of a book as a function of the number of words, rather than the fraction of the book, completed at any given point; we define this measure as ``cumulative word-time''.
Using ousiometrics, a reinterpretation of the valence-arousal-dominance framework of meaning obtained from semantic differentials, we convert text into time series of power and danger scores, with time corresponding to cumulative word-time.
Each time series is then decomposed using empirical mode decomposition into a sum of constituent oscillatory modes and a non-oscillatory trend.
By comparing the decomposition of the original power and danger time series with those derived from shuffled text, we find that shorter books exhibit only a general trend, while longer books have fluctuations in addition to the general trend.
These fluctuations typically have a period of a few thousand words regardless of the book length or library classification code, but vary depending on the content and structure of the book. 
Our findings suggest that, in the ousiometric sense, longer books are not expanded versions of shorter books, but rather are more similar in structure to a concatenation of shorter texts.
Further, they are consistent with editorial practices that require longer texts to be broken down into sections, such as chapters.
Our method also provides a data-driven denoising approach that works for texts of various lengths, in contrast to the more traditional approach of using large window sizes that may inadvertently smooth out relevant information, especially for shorter texts.
All together, these results open up avenues for future work in computational literary analysis, particularly the possibility of measuring a basic unit of narrative.
\end{abstract}

\maketitle

\section{Introduction}
\label{sec:intro}

The computational study of word usage in text has mainly been confined to aggregates, particularly the frequency of words in large corpora and contextual co-occurrence~\citep{corral_Zipf_2015, ryland_williams_zipfs_2015, dodds_allotaxonometry_2020, vaswani_attention_2017, reimers_SentenceBERT_2019}.
Only recently have there been quantitative studies on how word usage changes over the course of a single text, and these have focused on the structure of narratives. Studies on narratives have traditionally been made from a qualitative perspective, requiring human inputs to interpret text~\citep{freytag_freytags_1900, vonnegut_palm_1999,  ricoeur_narrative_1980, genette_narrative_1983, phelan_time_2012, brown_shapes_2020}. While the world’s literary experts are far from being supplanted by computational analyses---and arguably never will be---their attentions cannot be scaled to study very large corpora, for which quantitative techniques have been developed.  

Gao \etal~\citep{gao_multiscale_2016} showed that sentiment in novels have long-range correlations across the length of the book, indicating that sentiment exhibits a structure that relates to the flow of the novel.
Inspired by Vonnegut~\citep{vonnegut_palm_1999,vonnegut2005a,vonnegut2010a},
\citet{reagan_emotional_2016} examined how smoothed, ``distantly measured''~\citep{moretti2013a} happiness scores~\citep{dodds_temporal_2011} vary across the length of a book.
Reagan \etal\ find that these happiness time series across fiction books reveal six major emotional arcs, identified as: Rags-to-riches (rise), tragedy (fall), Icarus (rise-fall), Vonnegut's man-in-a-hole (fall-rise), Cinderella (rise-fall-rise), and Oedipus (fall-rise-fall). 
More recently,~\citet{boyd_narrative_2020} showed that the different parts of Freytag's dramatic arc~\citep{freytag_freytags_1900} can be differentiated by the dominant word usage. 
Articles and prepositions are heavily used in the exposition stage, where narrators establish the setting.
Once the reader has an understanding of the context of the story, plot progression can proceed with more pronouns and function words. 
As the story reaches a climax, more cognitive process words are used as the characters and narrator work through the conflict~\citep{boyd_narrative_2020}. 
While nonfiction works such as TED talks, US Supreme Court decisions and \textit{The New York Times} articles follow similar patterns in plot progression as works of fiction, they differ in patterns of staging and cognitive processes.

\citet{schmidt_plot_2015} put forward the idea of ``plot arcs'', where a text is seen as a path through topics that represent a multidimensional space.
Similarly,~\citet{toubia_how_2021} studied narratives as paths in high dimensional word embedding space. 
Dividing a text into segments represented by corresponding average word embedding vectors, the narrative is then described as the path in word embedding space obtained by moving along consecutive segments.
They identify three properties of these paths, namely the speed, volume and circuitousness in word embedding space, and examine how well these properties can predict the financial success of movies and TV shows as well as the number of citations in academic papers.

Though these studies look at changes in word usage across text, they focus on the general shape of narrative progression from beginning to end. 
Narrative is often characterized as a time series, with the general arc represented by the rise and fall of certain markers (such as happiness scores~\citep{reagan_emotional_2016}) throughout the plot in terms of the fraction of the book covered.
This normalization by the text length allows for a direct comparison of texts of different lengths, such as a short story and a novel, and the corresponding shapes of their respective arcs. 
However, this approach does not allow us to compare sections of a text with other sections of comparable length, for example, a chapter of a novel vs. a short story of similar length. 

To do this, not just for narratives but for any type of text, we have to use \textit{cumulative word-time}, or simply \textit{word-time}, which we define as the number, rather than the fraction, of words covered in a text at a given point.
Though it is numerically identical to the cumulative word count, we want to emphasize that while word count is often thought of as a static measure, word-time flows as a reader goes through a text.

However, research on word usage change in word-time is scant. To our knowledge, there has yet to be a quantitative study on how word usage changes in word-time across different texts.
We know from the prior studies mentioned earlier that word usage changes over the length of a book in a general manner.
Just as a larger narrative can be broken down into sub-narratives~\citep{wallace_multiple_2012}, can we find meaningful changes in word usage within sections of a book?
How do we characterize these changes?

\begin{figure}
    \centering
    \includegraphics[width=\columnwidth]{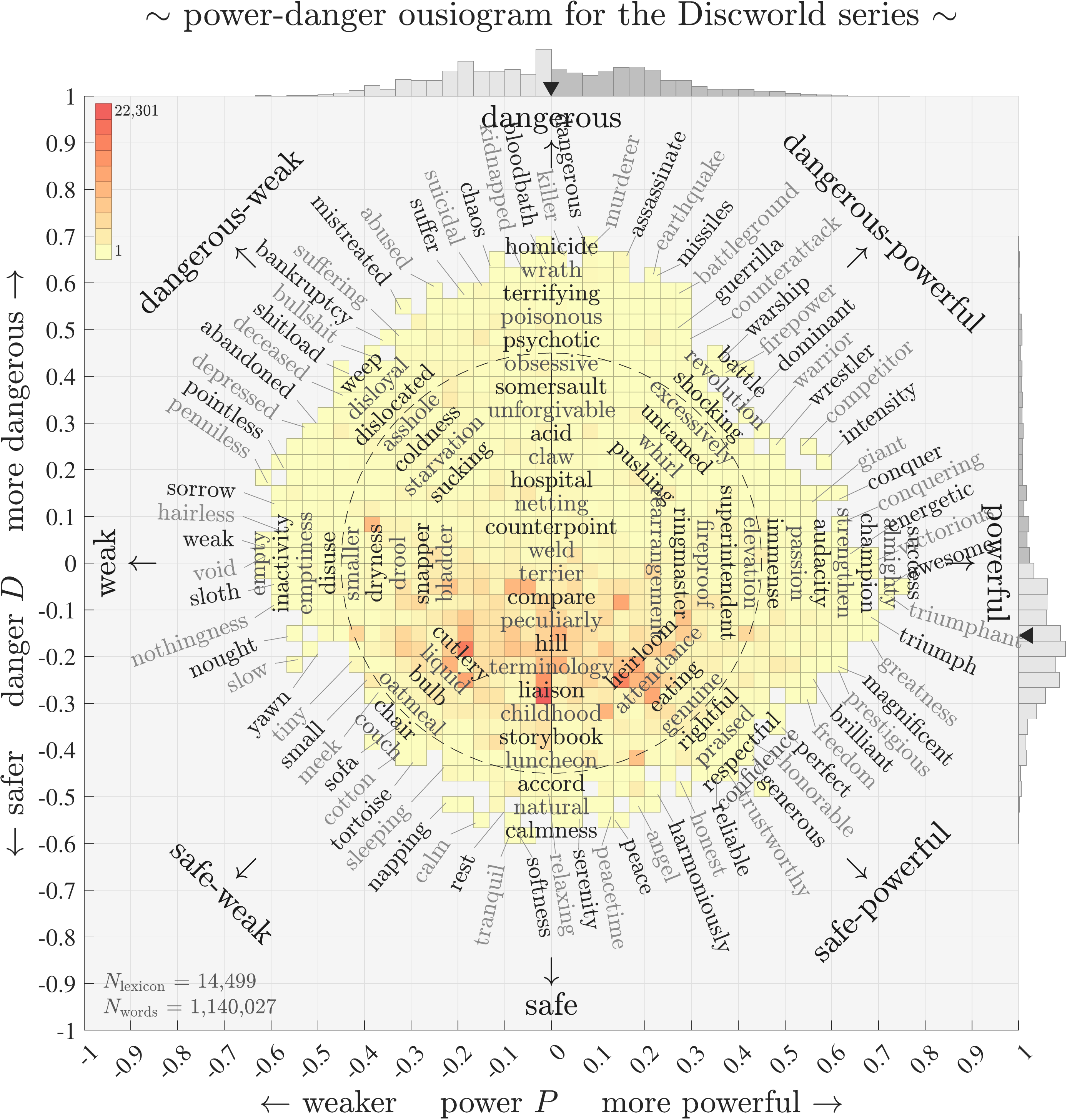}
    \caption{
    An `ousiogram'~\citep{dodds_ousiometrics_2021} displaying power and danger scores for a subset of 14,499 unique words appearing 
    in Terry Pratchett's 41-book Discworld series.
    The example words overlaid map the indicated eight major directions of the compass of essential meaning.
    The full set of 20,006 words with power and danger scores is derived from the NRC VAD lexicon~\citep{mohammad_obtaining_2018}.
    These 14,499 unique words collectively account for 1,140,027 total words (types versus tokens).
    The histogram's color map indicates frequency of usage. 
    The histogram, along with the marginal distributions for power and danger (right and top) present the same safety bias observed
    across disparate corpora~\citep{dodds_ousiometrics_2021}.
    }
    \label{fig:power_danger}
\end{figure}

While there are a number of ways to quantify word usage, we are particularly interested in how essential meaning changes over word-time. 
Since the 1950s, there have been efforts to distill the meaning of words into a few numbers. 
The valence-arousal-dominance (VAD) framework~\citep{osgood_measurement_1957} has been one of the most widely used systems in quantifying meaning through semantic differentials. 
This framework was developed using factor analysis of word scores provided by human annotators for a small set of words, which resulted in the three dimensions of valence, arousal and dominance.
However, efforts to create large VAD lexicons using human annotators~\citep{warriner_norms_2013,mohammad_obtaining_2018} have shown that valence, arousal, and dominance are linearly correlated when a larger set of words is used.
By transforming the NRC (National Research Council Canada) VAD lexicon---the largest VAD lexicon published with more than 20,000 curated words~\citep{mohammad_obtaining_2018}---via singular value decomposition,~\citet{dodds_ousiometrics_2021} showed that the VAD framework can be collapsed onto two linearly independent dimensions which align with the concepts of ``power'' and ``danger'', with the third linearly independent dimension (``structure'') being less relevant in real-world corpora than the former two.
Fig.~\ref{fig:power_danger} shows representative words in power-danger space for Terry Pratchett's Discworld series.
Real-world corpora also exhibit a bias towards low-danger words, which parallels the positivity bias in language~\citep{dodds_human_2015}.
Valence scores are highly correlated with happiness scores~\citep{dodds_ousiometrics_2021}, which have been used to quantitatively uncover emotional arcs in stories~\citep{reagan_emotional_2016}. 
Thus, we expect power-danger scores, both having a linear dependence on valence (see Supplementary Information), to also be suitable markers in quantifying structure in text and to give more information than using happiness scores alone, especially since the power-danger lexicon contains more words and would yield higher word coverage.

\begin{figure*}
    \centering
    \includegraphics[]{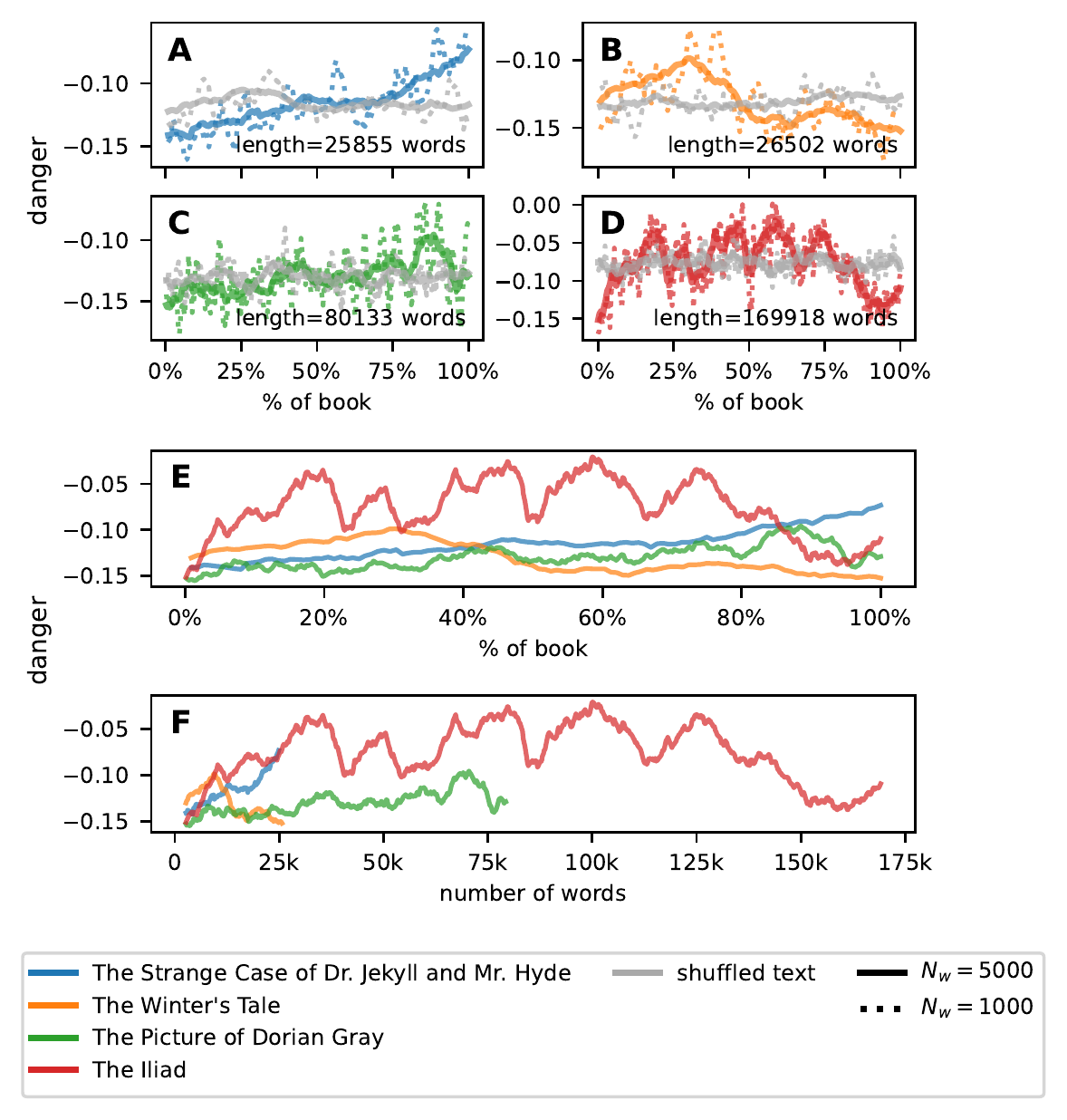}
    \caption{(a--d) These plots illustrate the effect of window sizes, shuffling, and book length on how danger scores change over the course of a book. Using larger window sizes (solid vs. dotted) results in smoother curves, although longer books tend to retain fluctuations in the time series for the same window size (a, b vs. c, d). Shuffling the text results in time series that are flatter than the original (gray). (e) Comparing books using normalized word-time shows more fluctuations for longer books than shorter ones, but (f) plotting the time series in raw word-time shows similarities between shorter books and sections of longer books. A skip size of $N_s=200$ words was used for all plots shown above.}
    \label{fig:time_series_summary}
\end{figure*}

An illustration of how a word usage marker, such as the danger score, changes over the course of a text is given in Figure~\ref{fig:time_series_summary}A--D. 
Each time series was constructed by splitting the text into windows of size $N_w$ words that skip every $N_s$ words; in previous studies, time series for all books were constructed using the same value for $N_w$, which range from a few hundred to a few thousand, with either $N_s$ or the the number of windows fixed.
Each window corresponds to a point in the time series and is characterized by the mean score of the words it contains.
Using a larger window size smooths out the time series, sometimes revealing the general shape, such as the steady increase in danger scores in ``The Strange Case of Dr. Jekyll and Mr. Hyde'' by Robert Louis Stevenson, or an up-down pattern as in ``The Winter's Tale'' by William Shakespeare.
Longer books, however, tend to retain fluctuations at the same window sizes.
If books of different lengths are compared by the fraction of the book covered, which we define as the \textit{normalized word-time} (Figure~\ref{fig:time_series_summary}E), it would seem that the word usage changes more steadily for shorter books than longer books.
However, if we compare the time series in (raw) word-time, so that the time series are of different lengths (Figure~\ref{fig:time_series_summary}F), the shorter time series appear comparable to \textit{sections} of the longer time series.
Thus, changes in word usage may be related to the length of a text, with longer texts having a similar structure to a concatenation of shorter texts.
In the context of narratives (as our examples in Figure~\ref{fig:time_series_summary} are), this suggests the possibility that long narratives may be composed of shorter ones, each with its own arc, that function as the basic unit of story.
Some of the fluctuations found in the time series may not be unwanted noise, but rather a measure of the lengths of these basic units.
Our aim is to characterize these fluctuations for various texts, and to relate them to properties of the texts themselves.

With this objective, we cannot use large window sizes, since they run the risk of smoothing out potentially relevant fluctuations.
However, we must also isolate the fluctuations that arise from noise, something we will likely see with smaller window sizes.
Since we are interested in how word usage changes over the course of a text, we can compare the time series to that obtained from a shuffled version of the text, which contains the same words but orders them randomly.
Using this as a reference isolates the effect of word order, and also avoids making assumptions on the nature of contaminating noise.
As expected, time series obtained from shuffled texts are flatter, with smaller fluctuations than those found in the original text (Figure~\ref{fig:time_series_summary}A--D).

To extract and characterize fluctuations in the time series at different scales, we use empirical mode decomposition (EMD), a technique that factors a signal into a sum of \textit{internal mode functions} (IMF), each of which is a mean-zero oscillatory time series with frequency and amplitude modulations, and a non-oscillating trend~\citep{huang_empirical_1998}.
While it is similar to wavelet decomposition in terms of its objective, EMD is data-adaptive, requiring almost no critical input from the user other than the raw time series itself, and is also well-suited for both nonstationarity and nonlinearity in time series.
A more detailed explanation of EMD is given in the Supplementary Information.

Figure~\ref{fig:sample_emd} shows the result of performing ensemble empirical decomposition (EEMD), an EMD variant that is more robust to noise~\citep{wu_ensemble_2009}, on a time series obtained from ``The Iliad''. While the original time series, derived from small, non-overlapping windows, contains significant noise, EEMD is able to separate the signal into components of different characteristic frequencies.
Further, we see that a partial reconstruction of the time series, obtained by summing the low-frequency IMFs, can replicate the time series obtained with larger window sizes, producing a denoised version.

\begin{figure*}
    \centering
    \includegraphics[]{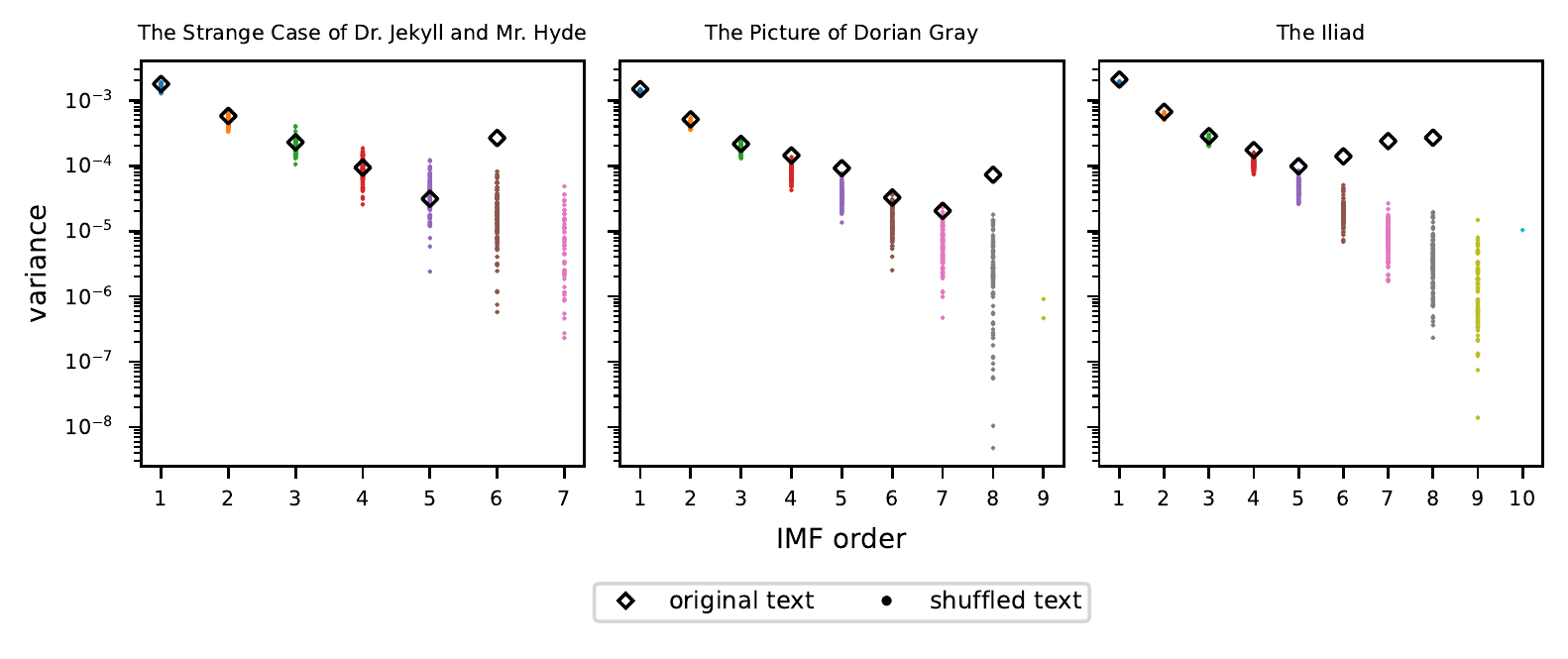}
    \caption{A comparison of the variances of the IMFs for the original text and 100 different realizations of shuffled text for three books in the dataset. Lower-order IMFs are more likely to correspond to noise, while higher-order IMFs are more likely to contain relevant information. Note that while technically the trend is not an IMF, for comparison purposes, the trend is included in this plot as the highest IMF order found for a given time series.}
    \label{fig:imf_variance_comparison}
\end{figure*}

We mentioned earlier that the time series for shuffled text are flatter than that of the original text (Figure~\ref{fig:time_series_summary}A--D), indicating that the IMFs of original and shuffled text may differ in their variances.
An illustration of how the variance changes as the IMF order increases for both the original and shuffled texts is given in Figure~\ref{fig:imf_variance_comparison}.
The variance of the IMFs in the shuffled texts generally decreases as the IMF order increases, similar to the observation for fractional Gaussian noise~\citep{wu_study_2004, flandrin_empirical_2004}.
However, this is not always true for the original texts.
For example, ``The Iliad'' shows a clear example of a book that differs in variance from the shuffled text from an IMF order below the trend, and continues to do so until the trend level.
On the other hand, ``The Picture of Dorian Gray'' clearly differs in the variance at an IMF order below the trend, but does not always do so for higher IMF orders.
In some books, such as ``The Strange Case of Dr. Jekyll and Mr. Hyde'', the original and shuffled versions only differ in the variance at the trend level.

To identify the cutoff IMF order for each book, we compare IMFs of the original text from those of different realizations of shuffled text.
The lowest IMF order at which the variance is higher than expected from the shuffled version is considered the cutoff at which word order becomes relevant.
Those where the cutoff order is an IMF (i.e., not the non-oscillating trend) are considered to have relevant fluctuations on top of the trend, while those that do not are considered to be trend-only.
To compare the variance, we use the method used by~\citet{wu_study_2004} and~\citet{flandrin_empirical_2004}, where the variances of the IMFs of the target series are rescaled so that the variances of the first IMFs of the target (original text) and reference (shuffled text) are comparable.
The assumption here is that the first IMF is noise; as we are using non-overlapping windows of size $N_w=50$, it is very difficult for a coherent narrative to be present at this scale, and the first IMF will most likely pick up noise due to the small window size.
Pairwise comparison of each IMF for the original and shuffled versions shows that the IMFs have similar periods for the first IMF, supporting this assumption.
We also verify that the periods are comparable up to the cutoff IMF (Figure~\ref{supp-fig:imf_ratios}).
While rescaling to the median of the first IMF is a reasonable choice, we also examine how rescaling to the 1st percentile of the first IMF or having no rescaling affects the results.

\begin{figure*}
    \centering
    \includegraphics[width=1.5\columnwidth]{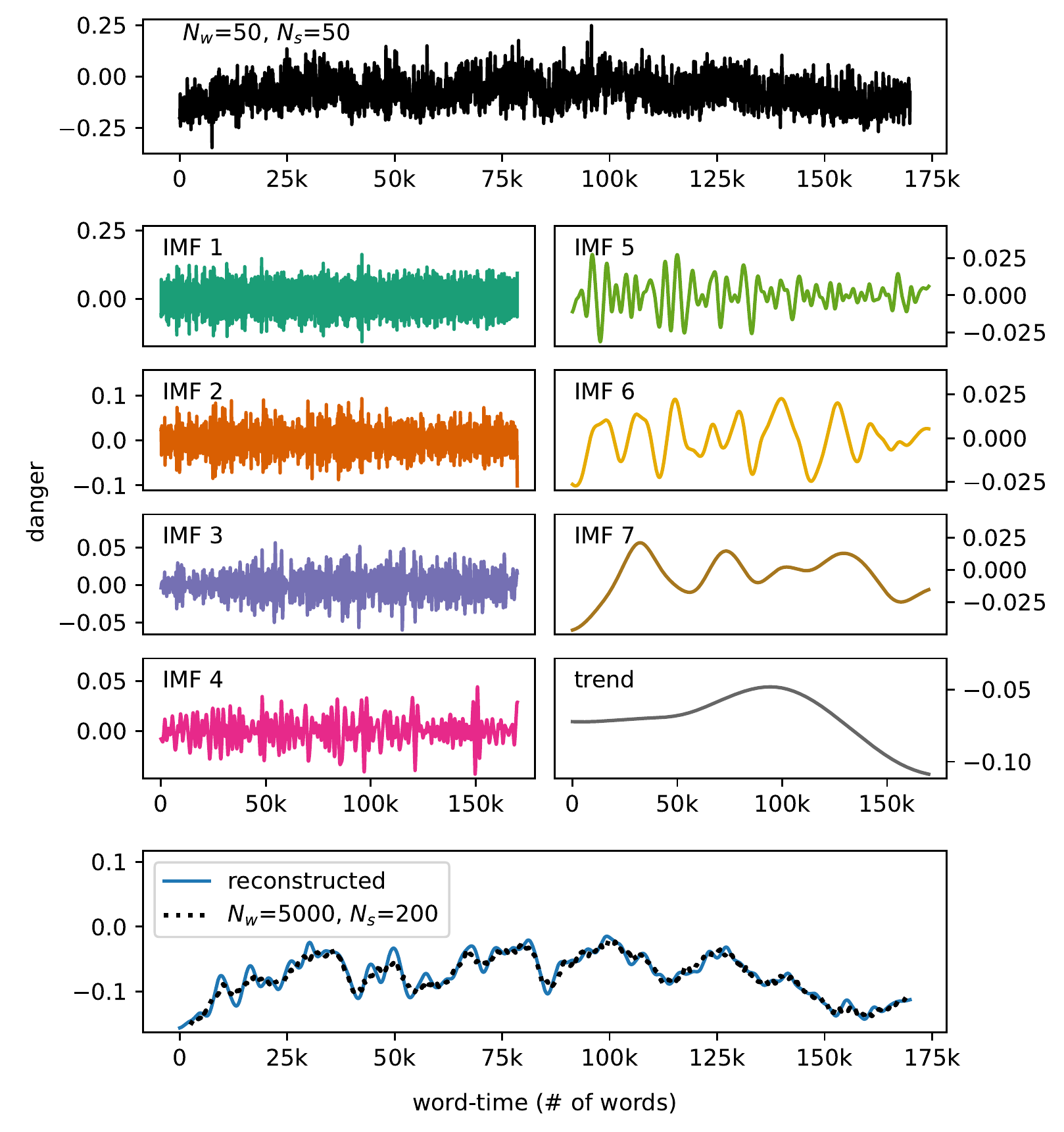}
    \caption{The decomposition of the danger time series of ``The Iliad'' using EEMD. The raw time series was obtained using non-overlapping windows with $N_w=50, N_s=50$ (top panel), and the IMFs and the trend obtained from EEMD are shown. The lowermost panel shows the sum of IMFs 5-7 and the trend (``reconstructed''), superimposed with the danger time series obtained using larger overlapping windows ($N_w=5000, N_s=200$). Note that the partial reconstruction from time series obtained using smaller windows is very similar to the raw time series obtained using larger, overlapping windows.}
    \label{fig:sample_emd}
\end{figure*}

While our study has been motivated by both qualitative and quantitative studies on narrative, we wish to study how word usage changes in word-time for a broader set of texts, both fiction and nonfiction, and discuss results for different categories as given by their Library of Congress Classification class and subclass labels.

\section{Methodology}
\label{sec:method}

\subsection{Data preprocessing}
We downloaded more than 45,000 books from Project Gutenberg~\citep{noauthor_project_nodate}, an online repository of books in public domain. The Gutenberg headers were removed using code from the Standardized Project Gutenberg Corpus~\citep{gerlach_standardized_2020}. Contractions, when unambiguous, were replaced with their expanded versions (e.g., ``n't'' to `` not''); if ambiguous, they were deleted, similar to what was done in~\citep{fudolig_sentiment_2022}. The remaining text was then converted to lowercase and tokenized using whitespace as separators, disregarding words that contain non-word characters and digits, and ignoring punctuation marks. This converts the text into a sequence of words.

\begin{figure}
    \centering
    \includegraphics[]{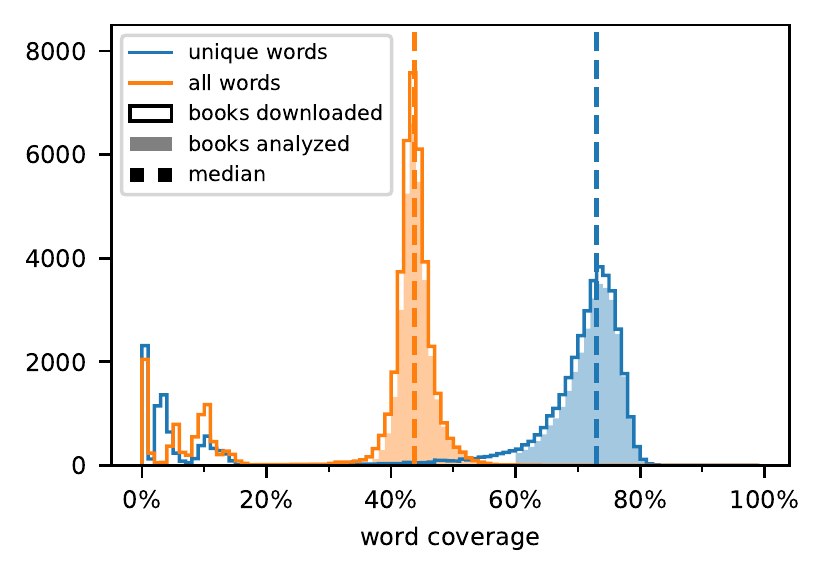}
    \caption{Word coverage of the PDS lexicon. The blue histograms correspond to the unique word coverage, while the orange histograms correspond to the word coverage allowing for word repetition. We show the histograms for all the books downloaded in our dataset (empty bars) as well as those we analyzed (shaded bars), and the dashed lines show the median word coverage of the books analyzed. The books included in our analysis do not include duplicate titles and must have (1) a unique word coverage of at least 60\%, (2) at least one lexicon word in every window and (3) intrinsic mode functions (IMFs) that can be successfully computed.}
    \label{fig:word_coverage}
\end{figure}

We then examine the word coverage of the power-danger lexicon. While the original NRC-VAD lexicon~\citep{mohammad_obtaining_2018}, from which the danger and power scores were derived, contained around 20,000 words, we expanded this to include noun plurals and conjugated forms not in the lexicon. Scores of the base forms of the verbs and nouns were used as the scores of their conjugated versions, expanding the lexicon to 32,721 words (the lexicon is available at \href{https://doi.org/10.5281/zenodo.7816312}{https://doi.org/10.5281/zenodo.7816312}). We only consider books with a 60\% unique word coverage, which consist almost exclusively of English text and cover 93\% of all English books in the downloaded set. The median unique word coverage of this subset of books is 73\%. Filtering for word coverage, removing books with duplicate titles, as well as requiring that the time series must have at least one lexicon word and that the ensemble empirical mode decomposition (EEMD; see Section~\ref{subsec:methods_relfluc}) can be successfully computed (i.e., the EEMD decomposes up to the trend level, such that the mean of the sum of the EEMD results is within 10\% of the mean of the original time series), leaves us with 31,690 books (Figure~\ref{fig:word_coverage}).

\subsection{Constructing the time series from text}
We construct the danger and power time series by segmenting the sequence of words into non-overlapping windows of size $N_w=50$, each of which corresponding to a point in the time series.
In each window, we take each word $w_i$ with score $s_i$ in the lexicon that occurs $n_i$ times in the window.
If there are $m$ unique words in the window that are in the lexicon, then the score for the window is

\begin{equation}
    s_w = \frac{\sum_{i=1}^m n_i s_i}{\sum_{i=1}^m n_i}
\label{eq:score}
\end{equation}

The reference time series are constructed by shuffling the tokenized version of the text, performing the windowing technique, and recomputing the average scores for each window.
This assures that the time series of both the target (original text) and the reference (shuffled text) are of the same length.

\subsection{Characterizing relevant fluctuations}
\label{subsec:methods_relfluc}

We use ensemble empirical mode decomposition (EEMD)~\citep{wu_ensemble_2009} with an ensemble size of 100 to obtain the internal mode functions (IMFs) for the time series corresponding to the original text.
Each time series in the ensemble is the sum of the raw time series and white noise with a standard deviation $0.2\sigma$, where $\sigma$ is the standard deviation of the raw time series, as suggested by~\citet{wu_ensemble_2009}.
As our reference time series, we use 100 different shuffled versions of the original text.
For each of the reference time series, we use basic empirical mode decomposition (EMD) to obtain the IMFs.
We use code from the Python package \texttt{emd}~\citep{quinn_emd_2021} to perform all EMD-related calculations.

We then rescale the variances of the IMFs of the time series as suggested in~\citep{wu_study_2004, flandrin_emd_2005}.
The variance for an IMF that has zero mean by definition is $\textrm{Var} = \sum_i{x^2_i} / N$, where $N$ is the length of the time series and $x_i$ is the value at word-time point $i$. If the variance of the IMF order $i$ in the target (original text) is $\textrm{Var}_{T,i}$ and $\textrm{Var}_{R, 1}$ is the representative value of the first IMF of the reference (shuffled text), then the rescaled variances are $\textrm{Var}^\prime_{T,i} = \textrm{Var}_{T,i} \frac{\textrm{Var}_{R,1}}{\textrm{Var}_{T, 1}}$.
In the log-scale plot given in Figure~\ref{fig:imf_variance_comparison}, this is equivalent to shifting the variance curve of the original text up or down so that the rescaled first IMF of the original text is equal to a representative value of the first IMFs of the shuffled texts ($\textrm{Var}^\prime_{T,1} = \textrm{Var}_{R,1}$).
Three different representative values were considered based on the distribution of the first IMFs of the reference time series: the median, the 1st percentile, and the variance of the first IMF of the target (no rescaling). The lowest IMF order at which the rescaled variance is higher than the 99th percentile of the variances for shuffled text is considered as the cutoff IMF order.

Once the cutoff IMF order is identified, the corresponding period is computed from the center of the frequency bin with the highest energy as obtained using the Hilbert-Huang transform (HHT).
Since we want to count periods in the unit of number of words, we compute for the HHT setting the sampling rate at ${N_s}^{-1} \textrm{word}^{-1}$, where $N_s=50$ is the skip size in the windowing procedure.
We use logarithmically spaced frequency bins, spanning from $10^{-6} \textrm{ word}^{-1}$ to $1 \textrm{ word}^{-1}$, resulting in a range of period values between 1 to $10^6$ words, chosen because none of the texts examined exceed $10^6$ words in length.
The choice of logarithmic spacing is motivated in part by how EMD on white noise performs like a dyadic filter, with IMF frequencies decreasing by roughly a factor of 2 for every order.
Further, in our preliminary analysis for select books, we find that logarithmic spacing provides an adequate representation of the spectra, especially since the IMF frequencies and periods span orders of magnitude.
We emphasize that obtaining a characteristic value for the \textit{period} of the IMF is independent of the method used to extract the cutoff IMF \textit{order} discussed earlier, and that the bins used for the HHT are the same across all texts.

\section{Results}
\label{sec:results}

We analyze more than 30,000 books from Project Gutenberg that passed our selection criteria.
The selected books are almost exclusively in English, with at least 60\% of the unique words in each book included in the lexicon.
Around 60\% of the books are in the ``Language and Literature'' Library of Congress Classification (LCC) code  (class label ``P''), while the remaining are spread out among the various LCC class labels, with ``World History'' (class label ``D'') as the next largest category of books in the dataset (around 8\% of the books).
While we performed the analysis for both danger and power scores, the results are similar for both, and we only discuss the danger scores in the main text. The results for power scores are included in the Supplementary Information.

\subsection{Cutoff IMF orders}

\begin{figure*}
    \centering
    \includegraphics[]{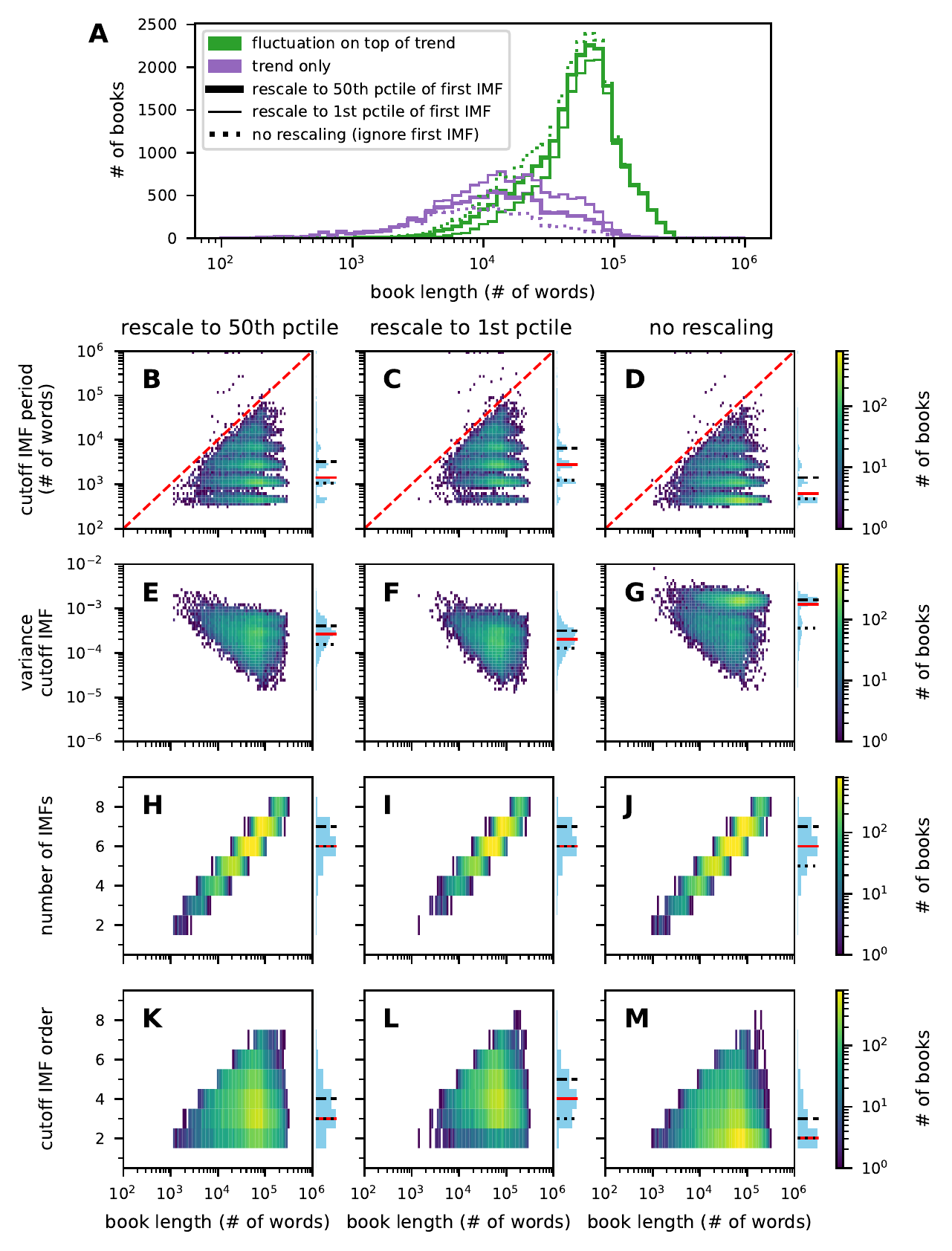}
    \caption{Characterizing relevant fluctuations in the danger time series of books in the corpus. (a) These histograms show the number of books that have fluctuations on top of the trend (green) and those that do not (purple). The different line widths and line styles correspond to the different rescaling factors: solid thick lines for using the median of the first IMF, solid thin lines for using the 1st percentile of the first IMF, and dotted lines for no rescaling. (b\textendash d). These are heatmaps that show the relationship between the period of the cutoff IMFs (as the number of words) and the book length for the various rescaling factors (shown above each plot). We can see that most of the points fall below the 45 degree line (red dashed line), indicating that the cutoff IMF period is less than than the book length for the vast majority of books. The histograms for the cutoff IMF period are shown on the right side of each plot, with the 25th, 50th, and 75th percentiles shown in dotted black, solid red, and dashed black lines, respectively. The rest of the figures are similar to (b\textendash d), but for different quantities: variance (e\textendash g), number of IMFs (h\textendash j; excludes the trend), and cutoff IMF order, with the first IMF counted as 1 (k\textendash m).}
    \label{fig:cutoff_imf_summary}
\end{figure*}

Figure~\ref{fig:cutoff_imf_summary}A shows histograms of the number of words found in books that have a cutoff IMF order below the trend, and those that do not.
While the choice of the rescaling factor influences the stringency of the cutoff criteria, they also offer an insight into the robustness of the results.

Rescaling to the 1st percentile of the first IMF requires a higher threshold to differentiate the IMFs, and thus more books are classified as trend-only. For those classified as having fluctuations on top of the trend, the cutoff periods obtained may be higher. On the other hand, no rescaling makes it more likely to see differences in the lower IMF orders, resulting in lower predicted cutoff periods.
We note that when no rescaling is performed, the cutoff IMF order for many of the books was the first IMF, which does not make much sense given that the window size of 50 words is relatively small and that the first IMF only corresponds to a period of around 100 words.
We also know from the IMF decomposition of fractional Gaussian noise~\citep{wu_study_2004, flandrin_emd_2005} that the first IMF has a different behavior compared to the higher-order IMFs.
Thus, for purposes of comparison, we disregard the first IMF in obtaining the cutoff IMF order when no rescaling is applied.

The predicted cutoff IMF order, including the case when it does not exist, differs for majority of the books examined.
However, we find some general results that hold regardless of the rescaling factor used.
For instance, we find that longer books tend to have relevant fluctuations on top of the trend (i.e., have a cutoff IMF order that is not the trend) while shorter books do not (Figure~\ref{fig:cutoff_imf_summary}A), especially for books with less than 3000 words or greater than 100,000 words.
Books with word counts in between these values may or may not have relevant fluctuations on top of the trend and the choice of the rescaling factor may influence the prediction for the cutoff IMF order.
When rescaling by the 50th percentile of the first IMF, the 25th to 75th percentiles range from roughly 1000 to 3200 words; when rescaling by the 1st percentile of the first IMF, this changes to around 1200 to 6400 words; and when no rescaling is used, this changes to around 500 to 1400 words.
In the case when relevant fluctuations on top of the trend are found, no clear relation between the cutoff IMF period and the book length is observed.
These can be seen in the heat maps and the corresponding histograms in Figure~\ref{fig:cutoff_imf_summary}B--D.
We also note that, with very few exceptions, the cutoff period is less than the book length.
The striated pattern in the heatmaps for the cutoff periods is due to the discreteness of the EEMD, which is observed even in white noise~\citep{wu_study_2004}, where the EEMD acts like a dyadic filter.
While these results were generated using all the books in the corpus, both fiction and nonfiction, we also analyze the different book classifications in the next subsection and include more details in the Supplementary Information.

Similarly, the raw variances of the cutoff IMFs also do not exhibit any relationship with the book length (Figure~\ref{fig:cutoff_imf_summary}E--G).
The range of the variance values is also wider in comparison to that observed for the first IMF (Figure~\ref{supp-fig:imf_ratios}E--F), indicating that while the first IMF likely corresponds to noise due to the windowing technique used for all of the books, the books generally start to differ from each other at higher IMF orders.
Further, while the number of IMFs found by EEMD for the original time series increases with the book length (Figure~\ref{fig:cutoff_imf_summary}H--J), the cutoff IMF order itself shows no such correlation (Figure~\ref{fig:cutoff_imf_summary}K--M).

We find similar observations from our analysis on power time series. Shorter books are more likely to be trend-only, while longer books have relevant fluctuations above the general trend. While the cutoff periods are not identical to those found in danger scores, they are of the same order of magnitude (see Supplementary Information, Figure~\ref{supp-fig:cutoff_imf_summary_power},  Table~\ref{supp-tab:period_numwords}).

\subsection{Relationship to book content}

\begin{figure*}
    \centering
    \includegraphics[]{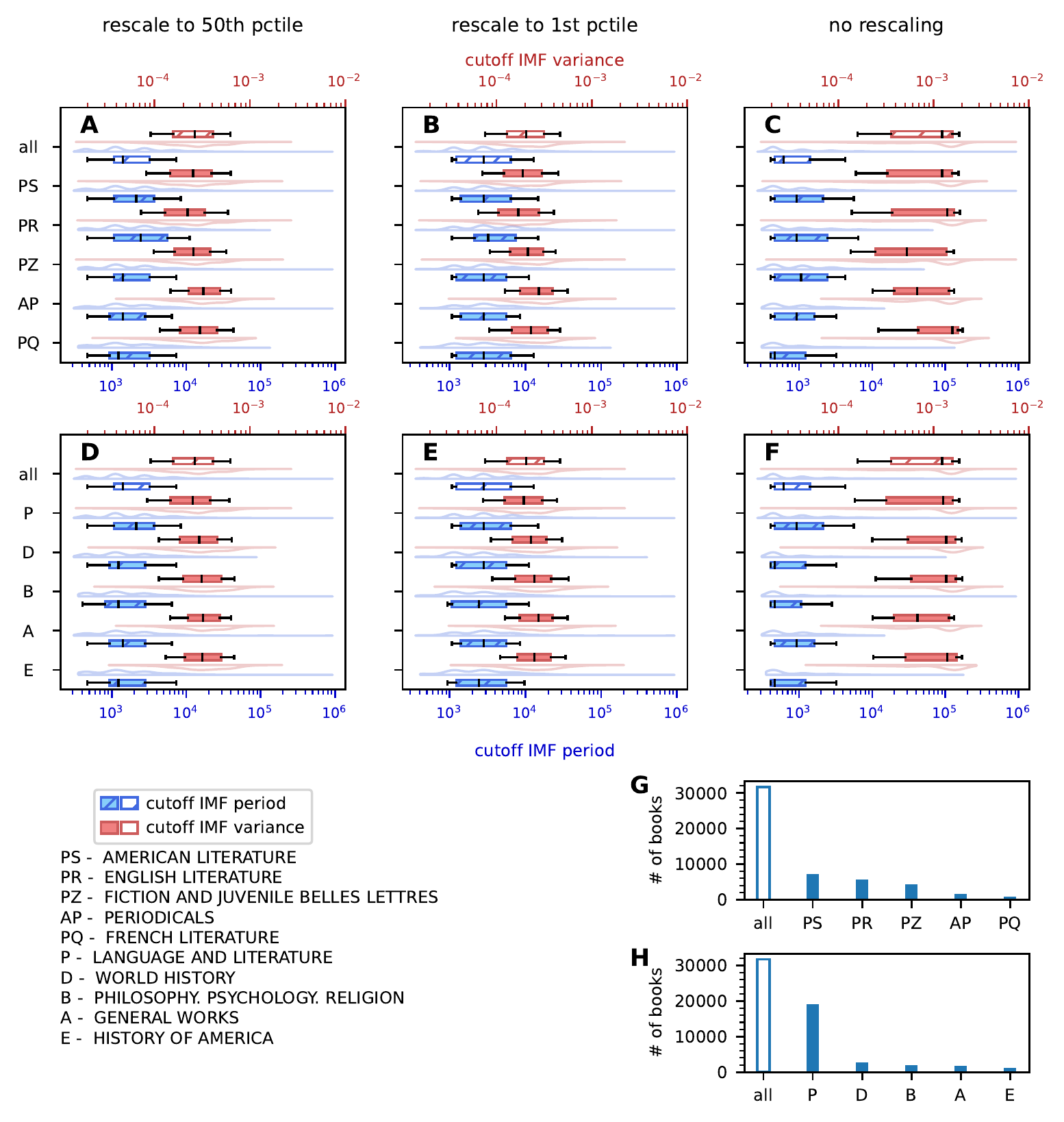}
    \caption{Periods and variances of the cutoff IMF in the danger time series for the top 5 LCC subclass and class labels with the most number of books in the dataset examined. (a-f) These are the boxplots for the periods (see x-axis at the bottom for scale) and the variances (see x-axis at the top for scale) of the cutoff IMF for books in the entire dataset (``all'') as well as that for each of the top 5 LCC subclass and class labels by size. The black line inside each box is the median, the box ranges from the 25th to 75th percentiles, and the whiskers extend from the 9th percentile to the 91st percentile. Violin plots are also included for more detail. (g-h) show the number of books analyzed in the dataset (``all'') and the top 5 subclass and class labels, including trend-only books.}
    \label{fig:boxplot_by_lcc}
\end{figure*}

The Gutenberg corpus assigns books to their Library of Congress Classification (LCC) subclass labels (e.g., PS for American Literature).
While it is possible for a book to have more than one LCC subclass label, around 95\% of the books we examined only had 1 label. The top 5 subclass labels with the most books in the dataset are American literature (PS), English literature (PR), Fiction and juvenile belles lettres (PZ), Periodicals (AP), and French/Italian/Spanish/Portuguese literature (PQ). We also looked at the class labels rather than the subclass labels (i.e., the first letter of the subclass labels).
Around 60\% of the class labels are for Language and Literature (P), while World History and History of Europe, Asia, Africa, Australia, New Zealand, etc. (D); Philosophy, Psychology and Religion (B); General Works (A); and History of America (E) round up the top 5 class labels.
While there are differences in the medians for the cutoff periods and variances across subclass and class labels, the spread of both the cutoff IMF period and variance are comparable across the top 5 class and subclass labels. This indicates that the breadth of the LCC system at the class or subclass levels obscures the differences that may arise from smaller groups within these categories (Figure~\ref{fig:boxplot_by_lcc}).

\begin{figure*}
    \centering
    \includegraphics[]{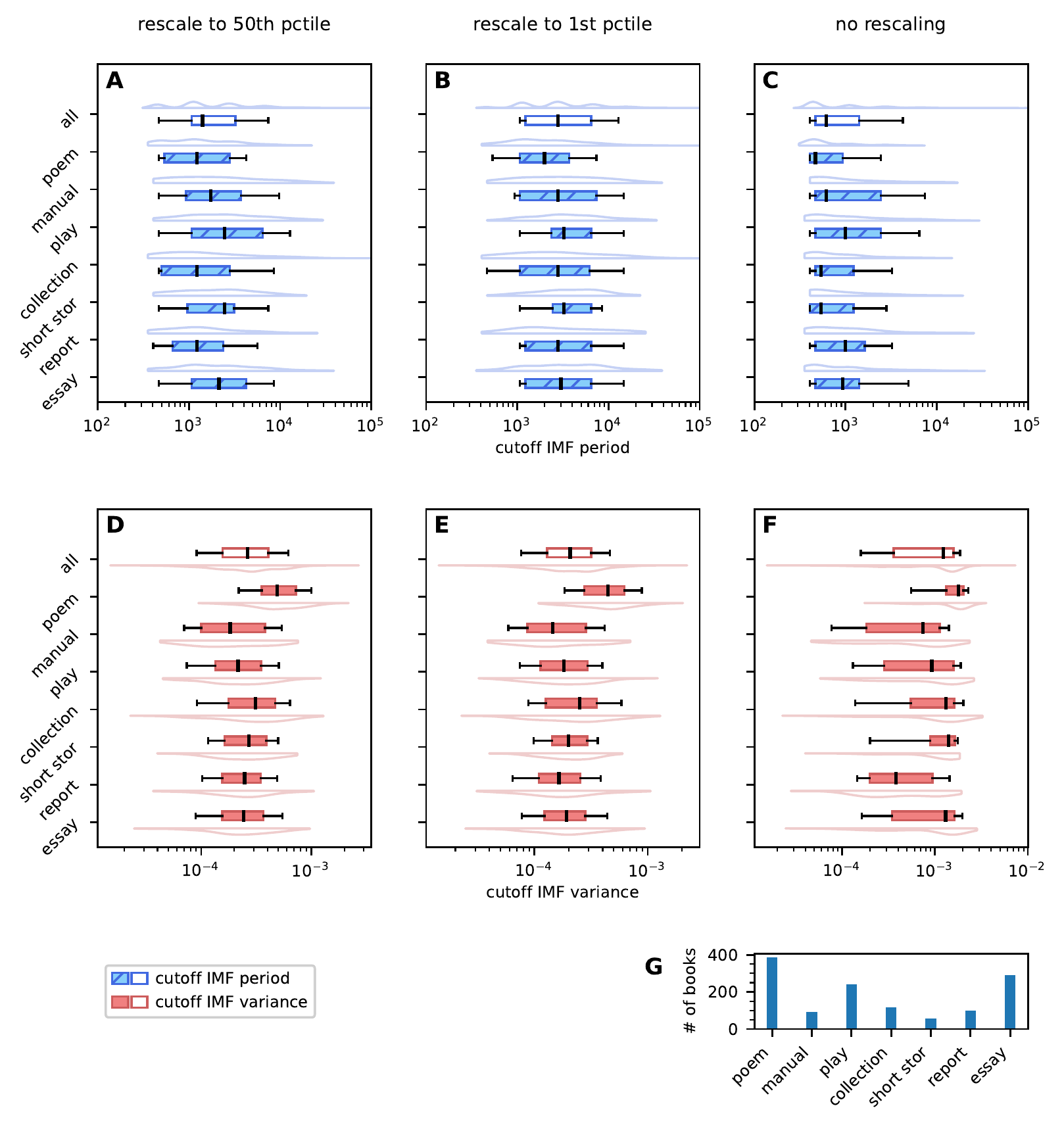}
    \caption{Periods and variances in the danger time series for books with a word in the title if they have a cutoff IMF order below the trend. Note that since the number of books for each word is much less than the total number of books examined, the boxplots for books not containing a given word in the title will be almost identical to the boxplots for the entire dataset (``all''). (a--c) show the boxplots for the cutoff IMF period, while (d--f) show the boxplots for the cutoff IMF variance for different rescaling factors. The black line inside each box is the median, the box ranges from the 25th to 75th percentiles, and the whiskers extend from the 9th percentile to the 91st percentile. Violin plots are also included for more detail. (d) shows the number of books with the given keywords in the title, including trend-only books.}
    \label{fig:words_in_title_boxplot}
\end{figure*}

On the other hand, we find that using very specific filters on the title of the book yields more insightful results (Figure~\ref{fig:words_in_title_boxplot}).
For books with a word beginning with ``poem'' in the title, those with a cutoff IMF order below the trend exhibit shorter cutoff IMF periods and markedly higher cutoff IMF variances compared to the rest of the dataset, which is consistent with poems being typically short, emotional and compact.
On the other hand, books with titles containing a word beginning with ``manual'' exhibit a lower median cutoff IMF variance.
This may be because books in this category (``A manual of clinical diagnosis'', ``The ladies' book of etiquette, and manual of politeness: a complete hand book for the use of the lady in polite society'', ``The skilful cook: a practical manual of modern experience,'' etc.) tend to be instructional and uniform in terms of topic and mood.
Books with words beginning with ``play'' have a higher median cutoff IMF period and lower cutoff IMF variance than books without.
Results for keywords such as ``collection'', ``short stor'' (includes both ``short story'' and ``short stories''), ``report'' and ``essay'' are more sensitive to the choice of rescaling factor.

For power time series, while the values for the cutoff IMF periods and variances obtained are different, we also find similar overlap across different LCC class and subclass labels. We also find differences across different books depending on words in their titles (see Supplementary Information, Figures~\ref{supp-fig:boxplot_by_lcc_power} and~\ref{supp-fig:words_in_title_boxplot_power}).

\section{Discussion}
\label{sec:summary}

\subsection{Summary of results}
We examine the fluctuations in the danger and power dimensions in ousiometrics, a reinterpretation of the valence-arousal-dominance framework, for more than 30,000 English books in Project Gutenberg.
Changes in word usage across text are analyzed by segmenting the text into windows and converting it to a time series.
While window size has conventionally been preset independent of the book length, we find that large window sizes remove oscillations in word usage in shorter books but retain them in longer books.
However, a visual examination of the time series in word-time, which we define as the number of words seen up to the present moment in a book, seems to indicate that the time series for shorter text are similar to subsections of longer text.
This suggests that in the ousiometric sense, longer texts are similar in structure to a concatenation of shorter texts, not unlike a novel broken into chapters.
We verify this by obtaining quantitative estimates of the periods of the relevant fluctuations in the time series.

We extract the different scales of fluctuations in the time series obtained from text using empirical mode decomposition (EMD).
EMD decomposes a time series into a non-oscillatory trend and a sequence of oscillatory intrinsic mode functions (IMFs) that differ in their characteristic frequencies.
It allows for both nonlinearity and nonstationarity, and gives an intuitive and data-driven understanding of the underlying fluctuations in a given time series.

For each book, we derive the danger and power time series for both the original text as well as an ensemble of shuffled versions.
By comparing the variances of the resulting IMFs of the time series obtained from the original text to those of the shuffled texts, we find that shorter books tend to exhibit only a non-oscillatory trend while longer books tend to exhibit relevant fluctuations with periods around the order of a few thousand words, shorter than the length of the book.
The period and variance of the relevant fluctuations do not depend on the book length, despite the number of IMFs increasing with the book length. Segregation of books by the Library of Congress Classification class or subclass codes does not result in well-defined groups that show variation in the periods and variances of the relevant fluctuations.
However, we observed differences when we applied very specific filters, such as words in a title.
We infer that what characterizes the scale of a book's relevant fluctuations is neither its general topic nor length, but rather more specific aspects such as its structure and content (e.g., poems vs. manuals).

The impact of our study is mainly on two fronts: (1) a quantitative analysis of the structure of texts as a function of word-time that is consistent with how longer texts are divided into meaningful sections, and (2) a method for denoising time series obtained from texts without resorting to using arbitrarily large window sizes.
We discuss these in the following paragraphs.

\subsection{Word counts and book types}
Segmenting longer texts into shorter, self-contained sections has long been acknowledged in the publishing world, as well as by researchers in the literature and computational domain~\citep{wallace_multiple_2012, reagan_emotional_2016}. 
While it is theoretically possible to write a basic narrative in a few words (e.g., ``For sale, baby shoes, never worn.'') or a hundred thousand, balancing the flexibility of longer texts with the constraints imposed by reader engagement and publication costs has given writers and editors rules of thumb for word counts.

Short stories and novellas are characterized by a focus on a single central conflict~\citep{masterclass_learn_2021} or a single chain of events~\citep{baldick_oxford_2015}.
In contrast, the longer word count of novels ($>$40,000 words~\citep{world_science_fiction_society_hugo_2022, science_fictions__fantasy_writers_of_america_nebula_2020}) allows for a fuller development of its characters and themes~\citep{baldick_oxford_2015} through the use of chapters, each of which is similar to a short story.
These are consistent with our results: trend-only books begin to decrease in number above a word count of 10,000, while the number of books with relevant fluctuations above the trend continues to increase, reaching its peak for books with 50,000 to 100,000 words (Figure~\ref{fig:cutoff_imf_summary}).
Further, the cutoff IMF periods are in the order of a few thousand words, comparable to the length of chapters, which are typically 1,500--5,000 words long~\citep{bingham_how_2020, noauthor_how_2017}.

While these are editorial guidelines for stories, we find that word usage in the ousiometric sense, through the power and danger time series, show segmentation for longer books across several categories, even those that do not necessarily qualify as literary narratives~\citep{piper_narrative_2021} (see Supplementary Information for a comparison between literature and non-literature books, Table~\ref{supp-tab:period_numwords}, Figures~\ref{supp-fig:cutoff_imf_summary_P}--\ref{supp-fig:cutoff_imf_summary_notP_power}).
Prior quantitative studies have shown that word usage patterns change over the course of text, not just for literary narratives but for other types of text (e.g., academic papers), albeit with different signatures~\citep{pechenick_characterizing_2015, boyd_narrative_2020}.
Segmentation in non-literary books is not uncommon as evidenced by the widespread use of chapters and sections, and is supported by our results.

Another result we wish to highlight is that the period of relevant fluctuations is independent of the book length.
As this period can be interpreted as the word count of meaningful segments, this suggests the existence of a basic unit of text of some characteristic length that serves as a building block to construct longer texts.
We suspect this may relate to the rate at which humans can process textual information, although testing this hypothesis is outside the scope of our work.

\subsection{Data-adaptive denoising of text-derived time series}
On the quantitative side, the method we used in this paper allows us to smooth out fluctuations in time series derived from text of various lengths.
While using large window sizes reduces noise, it is unclear if it also inadvertently smooths out relevant information, such as fluctuations associated with subplots. By performing partial reconstruction using the relevant IMF orders, we have a data-driven denoising approach that works for both short and long texts.

As our method relies heavily on empirical mode decomposition, it also carries the same limitations. 
Although EMD will ideally construct different modes for fluctuations of sufficiently different frequencies, mode mixing may occur due to signal intermittency. 
Using ensemble EMD (EEMD) mitigates this problem, but does not ensure that it will remove mode mixing in all cases.
We also note that we compare the original and shuffled texts only in the variance of their IMFs.
While we verified that the pairwise comparison of IMFs results in comparable IMF periods for the original and shuffled text up to the cutoff IMF order, we only defined a difference in IMFs in terms of their variance. While this method has been proposed for finding the appropriate cutoff IMF order in using EMD for denoising~\citep{wu_study_2004, flandrin_emd_2005}, it will miss any other difference that is not associated with the variance.
We also considered using the probability density function to compare IMFs; however, this method failed to produce accurate results in synthetic data, while the variance comparison method performed well in all the tests we performed. While our general observations are robust to the choice of parameters in the variance comparison method, different results may be obtained for a particular book depending on the parameters used. Thus, while our method can extract general trends on relevant fluctuations across a corpus, sensitivity analysis must be performed when results are to be obtained for a particular book.

\subsection{Future work}

As discussed earlier, our work opens up avenues to investigate building blocks of text in books; narratives would particularly be of interest.
In our study, we only look at the minimum word count at which word order becomes relevant in the ousiometric sense.
However, in addition to the cutoff IMF order from which this word count was obtained, the denoised ousiometric time series also includes contributions from higher IMF orders.
It would also be interesting to see whether there is a hierarchical structure in text segmentation.

We have studied the scale of fluctuations within texts in their danger and power time series.
While danger and power are orthogonal to each other and produce similar temporal results, we did not look at how they work together as a book progresses.
Similar to the work by~\citet{toubia_how_2021}, we hope to do a spatial analysis in power-danger space, specifically on the path taken by the narrative.
Other possible avenues for future work include expanding our corpus to include various texts, such as screenplays and movies, as well as comparing different versions of a book, such as the first draft and the final published version.

\section*{Ethical approval}
This article does not contain any studies with human participants performed by any of the authors.

\section*{Informed consent}
This article does not contain any studies with human participants performed by any of the authors.

\section*{Competing interests}
The authors declare no competing interests.

\section*{Acknowledgements}
The authors are grateful for the computing resources provided by the Vermont Advanced Computing Core and financial support from the Massachusetts Mutual Life Insurance Company. Computations were performed on the Vermont Advanced Computing Core supported in part by NSF award No. OAC-1827314. M.I.F. is also grateful to Miguel Fudolig, Joshua Minot, Andy Reagan and Aritra Banerjee for helpful discussions.

\section*{Data availability}
The source data is publicly available from Project Gutenberg~\citep{noauthor_project_nodate}. Instructions on how to generate and process the data used in the study are available at \href{https://doi.org/10.5281/zenodo.7816312}{https://doi.org/10.5281/zenodo.7816312}.




\clearpage

\end{document}


\title{\protect
Supplementary Information: A decomposition of book structure through ousiometric fluctuations in cumulative word-time
}

\maketitle


\subsection*{Power-danger framework}
While the valence-arousal-dominance (VAD) framework~\citep{osgood_measurement_1957} has been one of the longstanding theories for the quantification of meaning, attempts to measure VAD scores for a larger set of words reveal that these three dimensions are not necessarily orthogonal. In particular, the NRC VAD lexicon~\citep{mohammad_obtaining_2018}, which has around 20,000 words, shows moderate correlation between pairs of variables ($r_{V, A} \simeq -0.27$,  $r_{A, D} \simeq 0.30$, and  $r_{V, D} \simeq 0.49$). Performing singular value decomposition creates three new dimensions (goodness, energy and structure) that are linear combinations of valence, arousal and dominance, and a clockwise rotation of $\pi / 4$ in the goodness-energy plane creates a framework where each dimension can be interpreted by the words that occupy its extreme values~\citep{dodds_ousiometrics_2021}. The resulting three dimensions are power, danger, and structure, and are related to valence, arousal, and dominance in the following way:

\begin{equation}
    \begin{bmatrix}
    \textrm{Power} \\ \textrm{Danger} \\ \textrm{Structure}
    \end{bmatrix}
    = 
    \begin{bmatrix}
    \frac{1}{\sqrt{2}} &   \frac{1}{\sqrt{2}} & 0\\
    -  \frac{1}{\sqrt{2}} &   \frac{1}{\sqrt{2}} & 0 \\
    0 & 0 & 1 \\
    \end{bmatrix}
    \begin{bmatrix}
    +0.86 & -0.15 & +0.48 \\
    -0.16 & +0.83 & +0.54 \\
    +0.48 & +0.55 & -0.69
    \end{bmatrix}
    \begin{bmatrix}
    \textrm{Valence$^{\prime}$} \\ \textrm{Arousal$^{\prime}$} \\ \textrm{Dominance$^{\prime}$}
    \end{bmatrix}
\end{equation}

\noindent where the $^\prime$ symbol indicates that the valence, arousal, and dominance values are rescaled to lie in the range $[-\frac{1}{2}, \frac{1}{2}]$. In real-world corpora, power and danger are the more relevant dimensions. We refer the reader to~\citep{dodds_ousiometrics_2021} for an in-depth discussion of the power-danger-structure framework.

\clearpage
\subsection*{Comparison between first and cutoff IMFs}
\begin{figure*}[ht!]
    \centering
    \includegraphics[width=\textwidth]{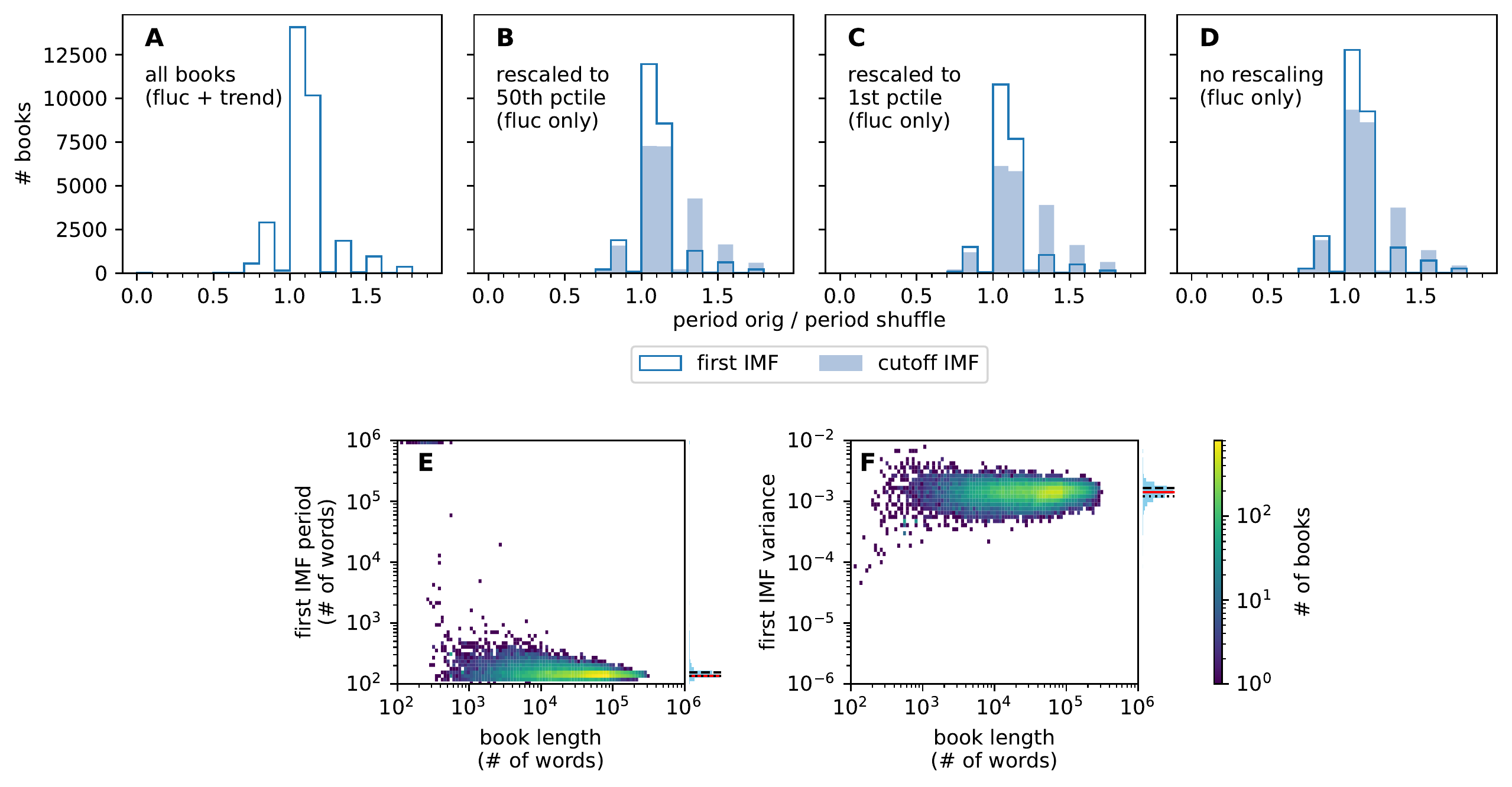}
    \caption{Histograms of the median ratios (measured across the different shuffling realizations) of the period of the IMF corresponding to the original text to that of the shuffled text for the \textit{danger} time series, as well as the period and variance corresponding to the first IMF. (a--d) The unshaded histograms refer to ratios for the first IMF, while the shaded histograms refer to the ratios for the cutoff IMF. In (a), we include all books, whether a cutoff IMF order below the trend was detected (``fluc'') or not (``trend''). In (b)--(d), we only include the ratios for the books where a cutoff IMF order below the trend was detected using the variance comparison method when the IMFs are rescaled by the rescaling factor (50th percentile for (b), 1st percentile for (c), no rescaling for (d)). (e--f) These heatmaps summarize the period and variance corresponding to the first IMF of the books in the dataset as a function of the book length. The right side of each heatmap shows the corresponding histogram and the 25th (black, dotted line), 50th (red, solid line) and 75th (black, dashed line) percentiles.}
    \label{fig:imf_ratios}
\end{figure*}

\begin{figure*}[ht!]
    \centering
    \includegraphics[width=\textwidth]{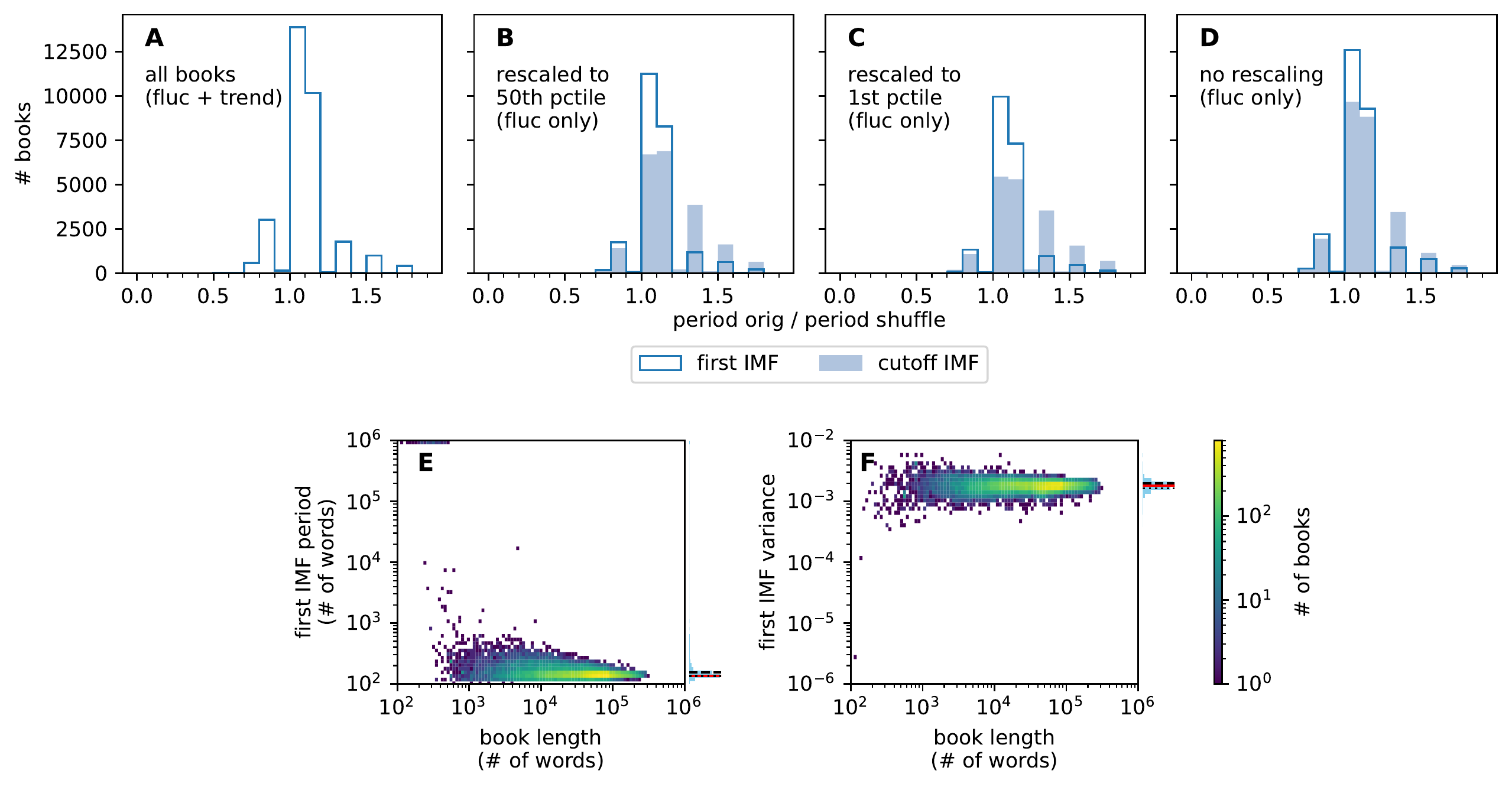}
    \caption{Histograms of the median ratios (measured across the different shuffling realizations) of the period of the IMF corresponding to the original text to that of the shuffled text for the \textit{power} time series, as well as the period and variance corresponding to the first IMF. (a--d) The unshaded histograms refer to ratios for the first IMF, while the shaded histograms refer to the ratios for the cutoff IMF. In (a), we include all books, whether a cutoff IMF order below the trend was detected (``fluc'') or not (``trend''). In (b)--(d), we only include the ratios for the books where a cutoff IMF order below the trend was detected using the variance comparison method when the IMFs are rescaled by the rescaling factor (50th percentile for (b), 1st percentile for (c), no rescaling for (d)). (e--f) These heatmaps summarize the period and variance corresponding to the first IMF of the books in the dataset as a function of the book length. The right side of each heatmap shows the corresponding histogram and the 25th (black, dotted line), 50th (red, solid line) and 75th (black, dashed line) percentiles.}
    \label{fig:imf_ratios_power}
\end{figure*}

\clearpage
\subsection*{Analysis of power time series}
\begin{figure*}[ht!]
    \centering
    \includegraphics[height=0.7\textheight]{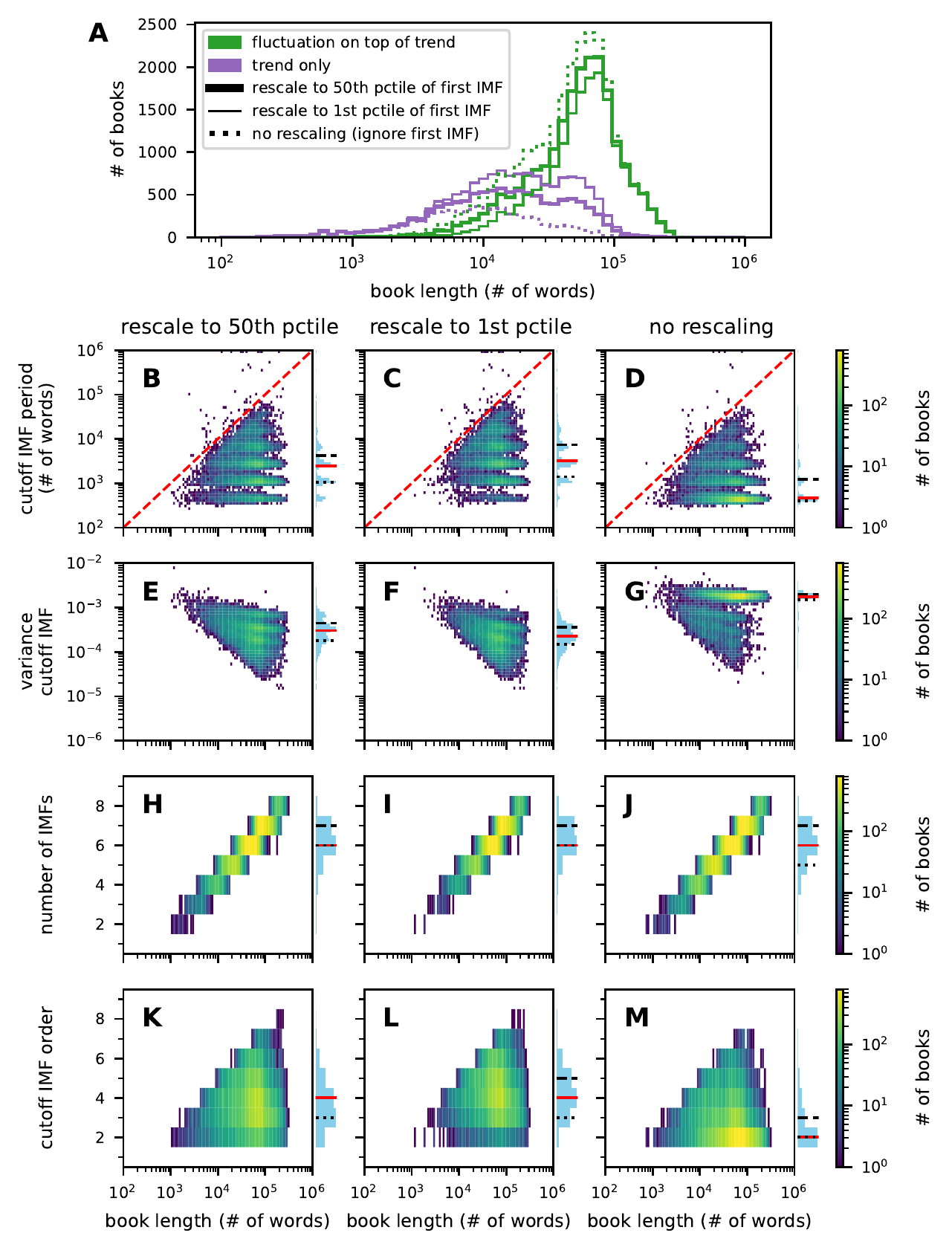}
    \caption{Characterizing relevant fluctuations in the power time series of books in the corpus (c.f. Figure~\ref{fig:cutoff_imf_summary}). (a) These histograms show the number of books that have fluctuations on top of the trend (green) and those that do not (purple). The different line widths and line styles correspond to the different rescaling factors: solid thick lines for using the median of the first IMF, solid thin lines for using the 1st percentile of the first IMF, and dotted lines for no rescaling. (b\textendash d). These are heatmaps that show the relationship between the period of the cutoff IMFs (as the number of words) and the book length for the various rescaling factors (shown above each plot). We can see that most of the points fall below the 45 degree line (red dashed line), indicating that the cutoff IMF period is less than than the book length for the vast majority of books. The histograms for the cutoff IMF period are shown on the right side of each plot, with the 25th, 50th, and 75th percentiles shown in dotted black, solid red, and dashed black lines, respectively. The rest of the figures are similar to (b\textendash d), but for different quantities: variance (e\textendash g), number of IMFs (h\textendash j), and cutoff IMF order, with the first IMF counted as 1 (k\textendash m).}
    \label{fig:cutoff_imf_summary_power}
\end{figure*}

\begin{figure*}[ht!]
    \centering
    \includegraphics[]{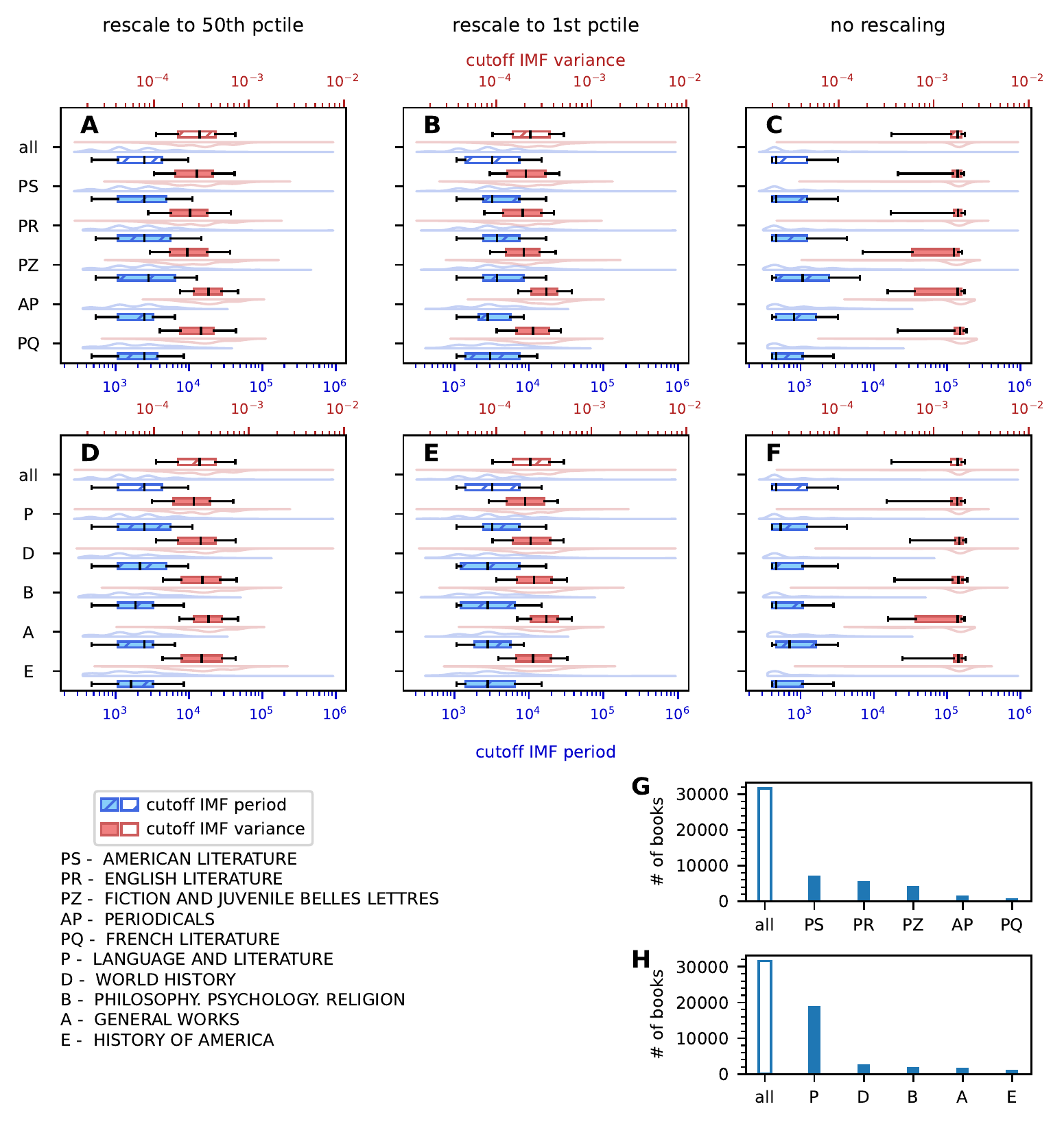}
    \caption{Periods and variances of the cutoff IMF in the power time series for the top 5 LCC subclass and class labels with the most number of books in the dataset examined. (a-f) These are the boxplots for the periods (see x-axis at the bottom for scale) and the variances (see x-axis at the top for scale) of the cutoff IMF for books in the entire dataset (``all'') as well as that for each of the top 5 LCC subclass labels by size. The black line inside each box is the median, the box ranges from the 25th to 75th percentiles, and the whiskers extend from the 9th percentile to the 91st percentile. Violin plots are also included for more detail. (g-h) shows the number of books analyzed in the dataset (``all'') and the top 5 subclass and class labels, including trend-only books.}
    \label{fig:boxplot_by_lcc_power}
\end{figure*}

\begin{figure*}[ht!]
    \centering
    \includegraphics[]{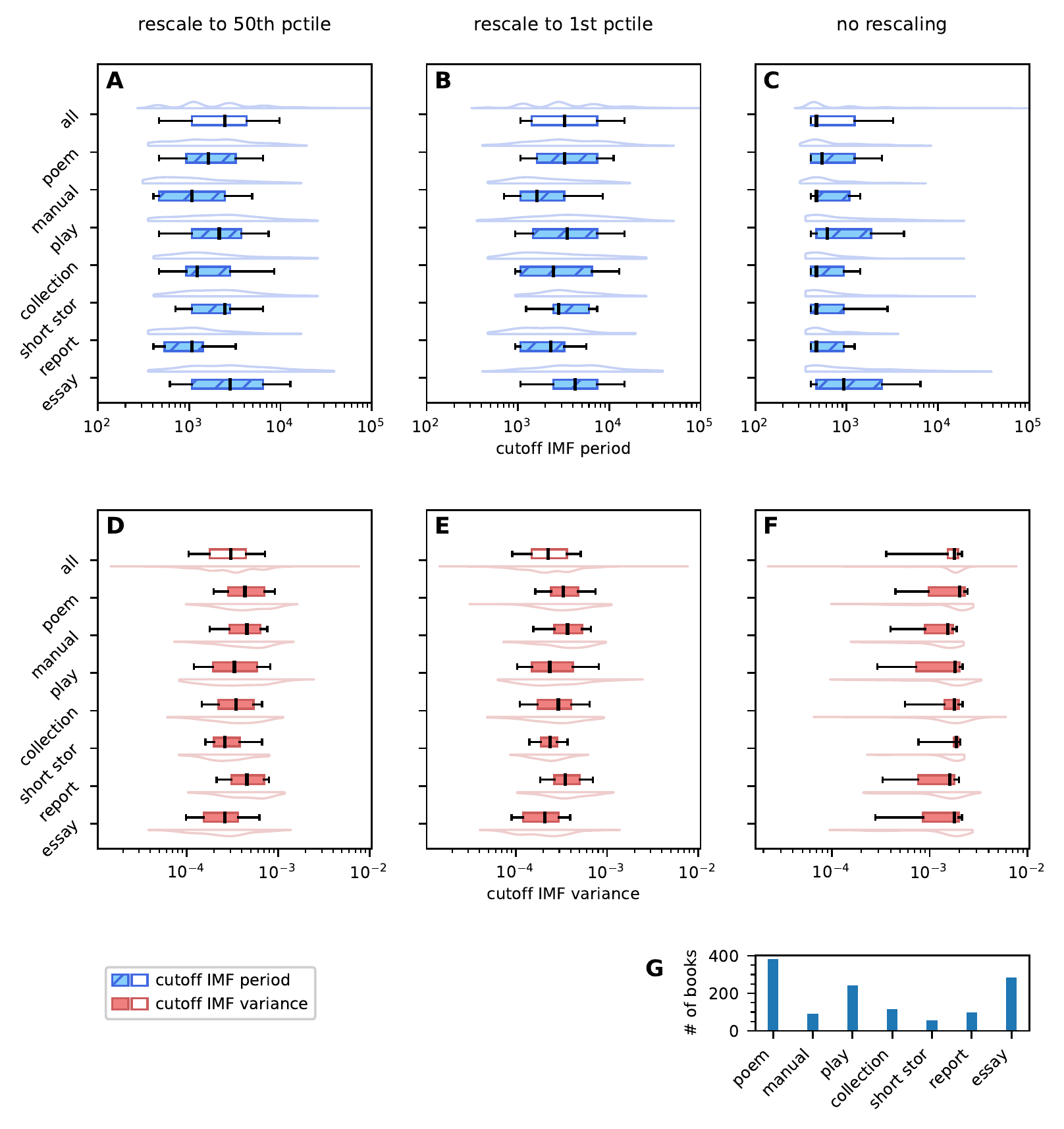}
    \caption{Periods and variances in the power time series for books with a word in the title if they have a cutoff IMF order below the trend. Note that since the number of books for each word is much less than the total number of books examined, the boxplots for books not containing a given word in the title will be almost identical to the boxplots for the entire dataset (``all''). (a--c) show the boxplots for the cutoff IMF period, while (d--f) show the boxplots for the cutoff IMF variance for different rescaling factors. The black line inside each box is the median, the box ranges from the 25th to 75th percentiles, and the whiskers extend from the 9th percentile to the 91st percentile. Violin plots are also included for more detail. (g) shows the number of books with the given keywords in the title, including trend-only books.}
    \label{fig:words_in_title_boxplot_power}
\end{figure*}

\clearpage
\subsection*{Comparison between literature and non-literature books}

\begin{table*}[ht!]
\begin{tabularx}{\textwidth}{lc|CCC|CCC|CCC}
\hline
 &
  \textit{Rescaling factor} &
  \multicolumn{3}{c|}{\textbf{50th percentile}} &
  \multicolumn{3}{c|}{\textbf{1st percentile}} &
  \multicolumn{3}{c}{\textbf{Without rescaling}} \\ \cline{3-11} 
 &
  \textit{Percentile} &
  \textbf{25th} &
  \textbf{50th} &
  \textbf{75th} &
  \textbf{25th} &
  \textbf{50th} &
  \textbf{75th} &
  \textbf{25th} &
  \textbf{50th} &
  \textbf{75th} \\ \hline
\multicolumn{1}{c}{\multirow{3}{*}{\rotatebox[origin=c]{90}{\textbf{Danger}}}} &
  All &
  1000 &
  1400 &
  3200 &
  1200 &
  2800 &
  6400 &
  500 &
  600 &
  1400 \\
  & Literature &
  1000 &
  2100 &
  3700 &
  1400 &
  2800 &
  6400 &
  400 &
  900 &
  2100 \\
  & Non-literature &
  900 &
  1200 &
  2800 &
  1200 &
  2800 &
  5600 &
  400 &
  500 &
  1200 \\ \hline
\multirow{3}{*}{\rotatebox[origin=c]{90}{\textbf{Power}}} &
  All &
  1000 &
  2400 &
  4200 &
  1400 &
  3200 &
  7400 &
  400 &
  500 &
  1200 \\
  & Literature &
  1000 &
  2400 &
  5600 &
  2400 &
  3200 &
  7400 &
  400 &
  500 &
  1200 \\
  & Non-literature &
  1000 &
  1400 &
  3200 &
  1200 &
  2800 &
  6400 &
  400 &
  500 &
  1100 \\ \hline
\end{tabularx}
\caption{Cutoff IMF period in number of words, rounded up to the nearest hundred, for danger and power time series for books with a cutoff IMF order below the trend.}
\label{tab:period_numwords}
\end{table*}

\begin{figure*}[ht!]
    \centering
    \includegraphics[]{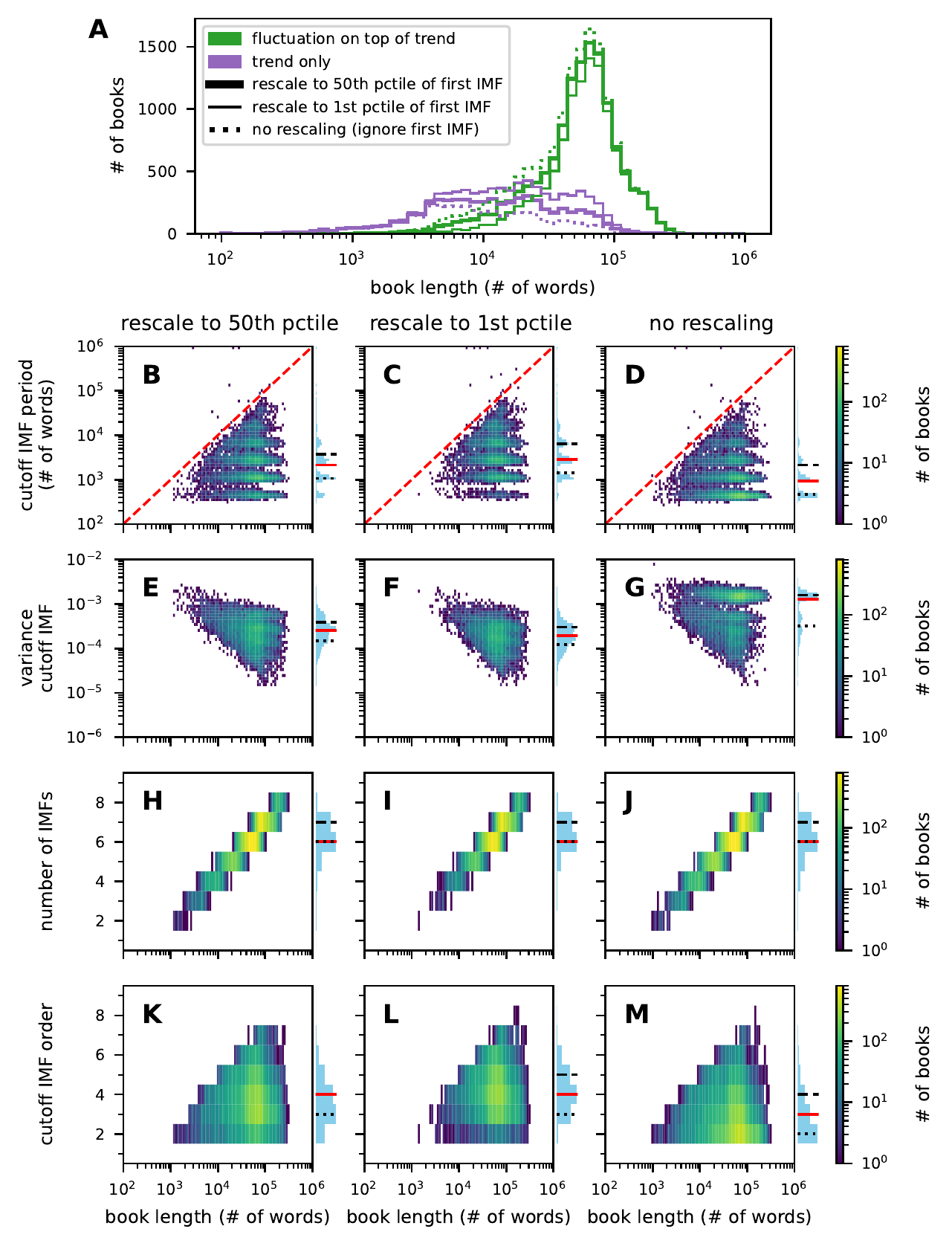}
    \caption{Characterizing relevant fluctuations in the \textit{danger} time series of books in the corpus for books under the Language and Literature Library of Congress Classification code ``P'' (c.f. Figure~\ref{fig:cutoff_imf_summary}). (a) These histograms show the number of books that have fluctuations on top of the trend (green) and those that do not (purple). The different line widths and line styles correspond to the different rescaling factors: solid thick lines for using the median of the first IMF, solid thin lines for using the 1st percentile of the first IMF, and dotted lines for no rescaling. (b\textendash d). These are heatmaps that show the relationship between the period of the cutoff IMFs (as the number of words) and the book length for the various rescaling factors (shown above each plot). We can see that most of the points fall below the 45 degree line (red dashed line), indicating that the cutoff IMF period is less than than the book length for the vast majority of books. The histograms for the cutoff IMF period are shown on the right side of each plot, with the 25th, 50th, and 75th percentiles shown in dotted black, solid red, and dashed black lines, respectively.The rest of the figures are similar to (b\textendash d), but for different quantities: variance (e\textendash g), number of IMFs (h\textendash j), and cutoff IMF order, with the first IMF counted as 1 (k\textendash m).}
    \label{fig:cutoff_imf_summary_P}
\end{figure*}

\begin{figure*}[ht!]
    \centering
    \includegraphics[]{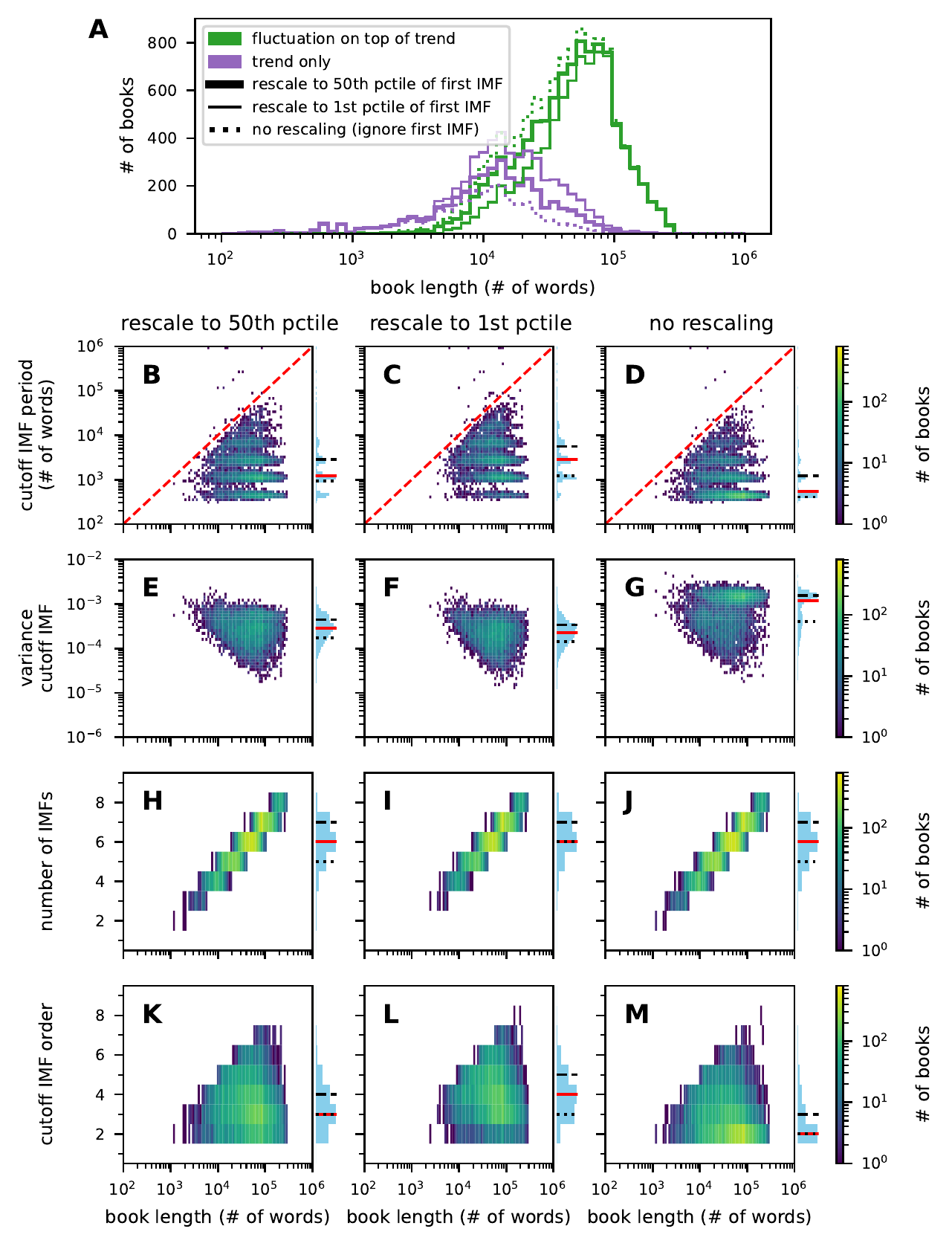}
    \caption{Characterizing relevant fluctuations in the \textit{danger} time series of books in the corpus for books in codes \textit{other than} the Language and Literature Library of Congress Classification code ``P'' (c.f. Figure~\ref{fig:cutoff_imf_summary}). (a) These histograms show the number of books that have fluctuations on top of the trend (green) and those that do not (purple). The different line widths and line styles correspond to the different rescaling factors: solid thick lines for using the median of the first IMF, solid thin lines for using the 1st percentile of the first IMF, and dotted lines for no rescaling. (b\textendash d). These are heatmaps that show the relationship between the period of the cutoff IMFs (as the number of words) and the book length for the various rescaling factors (shown above each plot). We can see that most of the points fall below the 45 degree line (red dashed line), indicating that the cutoff IMF period is less than than the book length for the vast majority of books. The histograms for the cutoff IMF period are shown on the right side of each plot, with the 25th, 50th, and 75th percentiles shown in dotted black, solid red, and dashed black lines, respectively.The rest of the figures are similar to (b\textendash d), but for different quantities: variance (e\textendash g), number of IMFs (h\textendash j), and cutoff IMF order, with the first IMF counted as 1 (k\textendash m).}
    \label{fig:cutoff_imf_summary_notP}
\end{figure*}

\begin{figure*}[ht!]
    \centering
    \includegraphics[]{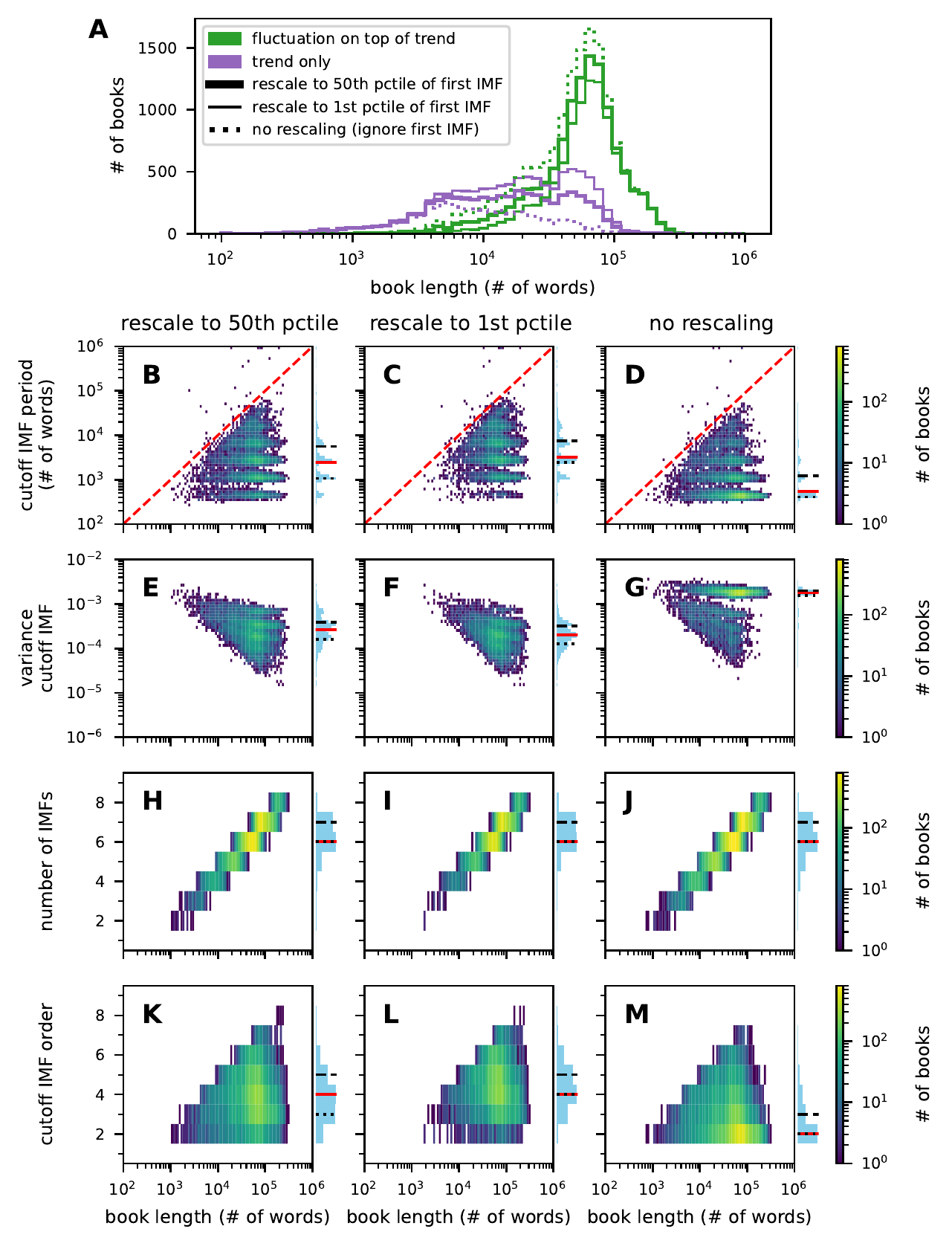}
    \caption{Characterizing relevant fluctuations in the \textit{power} time series of books in the corpus for books under the Language and Literature Library of Congress Classification code ``P'' (c.f. Figure~\ref{fig:cutoff_imf_summary}). (a) These histograms show the number of books that have fluctuations on top of the trend (green) and those that do not (purple). The different line widths and line styles correspond to the different rescaling factors: solid thick lines for using the median of the first IMF, solid thin lines for using the 1st percentile of the first IMF, and dotted lines for no rescaling. (b\textendash d). These are heatmaps that show the relationship between the period of the cutoff IMFs (as the number of words) and the book length for the various rescaling factors (shown above each plot). We can see that most of the points fall below the 45 degree line (red dashed line), indicating that the cutoff IMF period is less than than the book length for the vast majority of books. The histograms for the cutoff IMF period are shown on the right side of each plot, with the 25th, 50th, and 75th percentiles shown in dotted black, solid red, and dashed black lines, respectively.The rest of the figures are similar to (b\textendash d), but for different quantities: variance (e\textendash g), number of IMFs (h\textendash j), and cutoff IMF order, with the first IMF counted as 1 (k\textendash m).}
    \label{fig:cutoff_imf_summary_P_power}
\end{figure*}

\begin{figure*}[ht!]
    \centering
    \includegraphics[]{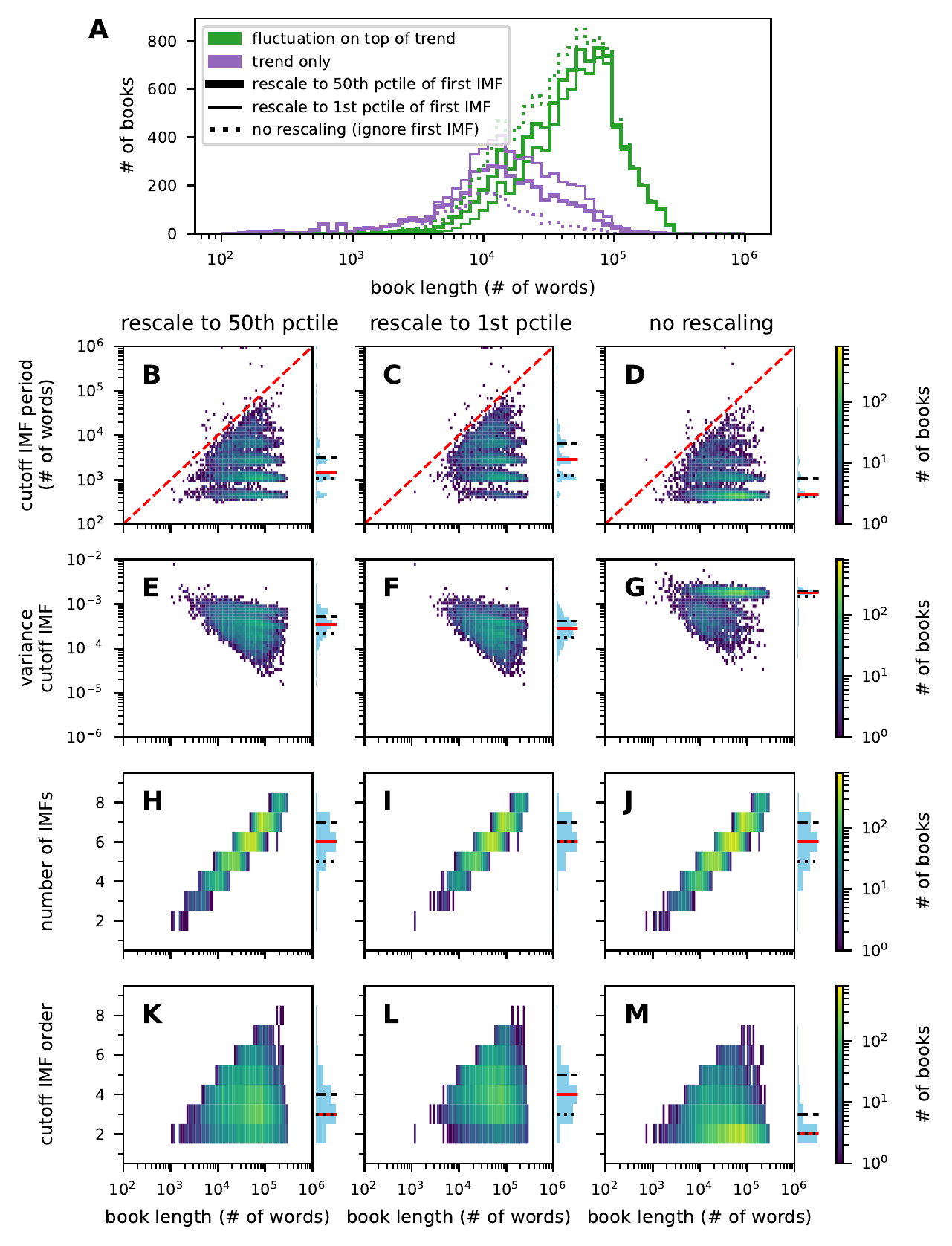}
    \caption{Characterizing relevant fluctuations in the \textit{power} time series of books in the corpus for books in codes \textit{other than} the Language and Literature Library of Congress Classification code ``P'' (c.f. Figure~\ref{fig:cutoff_imf_summary}). (a) These histograms show the number of books that have fluctuations on top of the trend (green) and those that do not (purple). The different line widths and line styles correspond to the different rescaling factors: solid thick lines for using the median of the first IMF, solid thin lines for using the 1st percentile of the first IMF, and dotted lines for no rescaling. (b\textendash d). These are heatmaps that show the relationship between the period of the cutoff IMFs (as the number of words) and the book length for the various rescaling factors (shown above each plot). We can see that most of the points fall below the 45 degree line (red dashed line), indicating that the cutoff IMF period is less than than the book length for the vast majority of books. The histograms for the cutoff IMF period are shown on the right side of each plot, with the 25th, 50th, and 75th percentiles shown in dotted black, solid red, and dashed black lines, respectively.The rest of the figures are similar to (b\textendash d), but for different quantities: variance (e\textendash g), number of IMFs (h\textendash j), and cutoff IMF order, with the first IMF counted as 1 (k\textendash m).}
    \label{fig:cutoff_imf_summary_notP_power}
\end{figure*}

\clearpage
\subsection*{Empirical mode decomposition}
Given a time series $x(t)$, EMD takes the mean $m_1$ of the upper envelope (connecting the local maxima by a cubic spline) and the lower envelope (connecting the local minima, also by a cubic spline).
The difference $h_{1, 1} = x(t) - m_{1, 1}$ is then checked for two conditions: (a) the number of extrema and the number of zero crossings must either equal or differ at most by one, and (b) the mean of $h_1$ must be zero.
If either condition is not met, this process is repeated $k$ times until both conditions are met, resulting in:
\begin{equation}
    h_{1, k} = h_{1, k-1} - m_{1, k}
\end{equation}

This process is called \textit{sifting}, and the function $c_1 := h_{1, k}$ is defined as the first IMF.
The difference $h_{2, 1} = x(t) - c_1$ is obtained and the same procedure is repeated, with each subsequent IMF corresponding to decreasing frequencies as the iteration progresses.
This iterative process continues $M$ times until $x(t) - c_M$ can no longer satisfy the conditions for an IMF; this is called the residual, $r_M$.
Thus, the original signal can be decomposed as the sum of the IMFs and the residual
\begin{equation}
    x(t) = \sum_{j=1}^M c_j + r_M .
\end{equation}
Each IMF can then be characterized by its frequency spectrum, which is obtained using the Hilbert (Hilbert-Huang) transform, and the characteristic frequency can be taken to be the center of the frequency bin with the highest energy.
The sum of the lower-frequency IMFs can then be used to denoise a time series. While similar to using a low-pass filter, using EMD also separates the fluctuations corresponding to each characteristic frequency, allowing comparison of IMFs across different temporal scales.

A commonly used variant is the \textit{ensemble empirical decomposition method} (EEMD), which was developed to reduce the effect of noise on the IMFs. Several realizations of white noise are added to the raw signal, and EMD is performed on each.
The EEMD then takes the average of the IMFs obtained across all these augmented signals.
Because of the ensemble averaging step, the resulting IMFs are more robust to the effect of noise.

While EMD can extract oscillatory modes of different frequencies from a time series, the method does not dictate which of the IMFs are due to noise and which are relevant to the true signal.
Methods developed include looking at the variances of the IMFs to set a cutoff IMF order~\citep{boudraa_emd-based_2007, boudraa_noise_2007}, comparing the correlation coefficients of each IMF and the raw time series~\citep{ayenu-prah_criterion_2010}, and comparing the probability density functions of the IMFs to that of the raw time series~\citep{komaty_emd-based_2014}. However, preliminary analysis showed that methods comparing the IMFs to the raw time series do not work sufficiently on our data, likely due to the choice of window size resulting in a low signal-to-noise ratio, so that the assumption that the relevant IMFs are closer to the original time series than the noise IMFs does not hold. We thus adapt a method that compares the time series to a null reference~\citep{wu_study_2004,flandrin_empirical_2004}, but use shuffled text instead of Gaussian noise as the null model.